%% file: main.tex
\documentclass[10pt,twocolumn,letterpaper]{article}
\PassOptionsToPackage{table}{xcolor}

\usepackage[pagenumbers]{cvpr} 

\input{preamble}

\definecolor{cvprblue}{rgb}{0.21,0.49,0.74}
\usepackage[pagebackref,breaklinks,colorlinks,allcolors=cvprblue]{hyperref}

\def\confName{CVPR}
\def\confYear{2027}

\title{GenLPO: Generation-Informed Local Preference Optimization for Driving World Models}

\author{%
    \parbox[t]{\dimexpr\textwidth-2\tabcolsep\relax}{%
        \centering\normalfont\large
        \mbox{Xunheng Kuang\textsuperscript{1,2,3,*}}\qquad
        \mbox{Zongshun Tian\textsuperscript{4,*}}\qquad
        \mbox{Mengfan Yuan\textsuperscript{5}}\qquad
        \mbox{Congqi Cao\textsuperscript{3,\textdagger}}\\[0.15em]
        \mbox{Yating Yu\textsuperscript{3}}\qquad
        \mbox{Changfeng Liu\textsuperscript{4}}\qquad
        \mbox{Lei He\textsuperscript{1,2,\textdagger}}\\[0.6em]
        \small
        \textsuperscript{1}School of Vehicle and Mobility, Tsinghua University, Beijing, 100084, China\\
        \textsuperscript{2}State Key Laboratory of Intelligent Green Vehicle and Mobility,
        Tsinghua University, Beijing, 100084, China\\
        \textsuperscript{3}School of Computer Science, Northwestern Polytechnical University, Xi'an Shaanxi, 710129, China\\
        \textsuperscript{4}State Key Laboratory of Intelligent Vehicle Safety Technology,
        Changan Automobile, Chongqing, 401133, China\\
        \textsuperscript{5}School of Mathematics and Statistics, Beijing Institute of Technology, Beijing, 100081, China\\[0.35em]
        \mbox{xunhengkuang@mail.nwpu.edu.cn}\qquad
        \mbox{congqi.cao@nwpu.edu.cn}\qquad
        \mbox{helei2023@tsinghua.edu.cn}\\[0.4em]
        {\fontfamily{cmtt}\selectfont\urlstyle{same}%
        \href{https://xunhengkuang.github.io/genlpo/}{\nolinkurl{https://xunhengkuang.github.io/genlpo}}}
    }%
}

\begin{document}
\twocolumn[{%
    \maketitle
    \noindent\begin{minipage}{\textwidth}
        \centering
        \includegraphics[width=\textwidth]{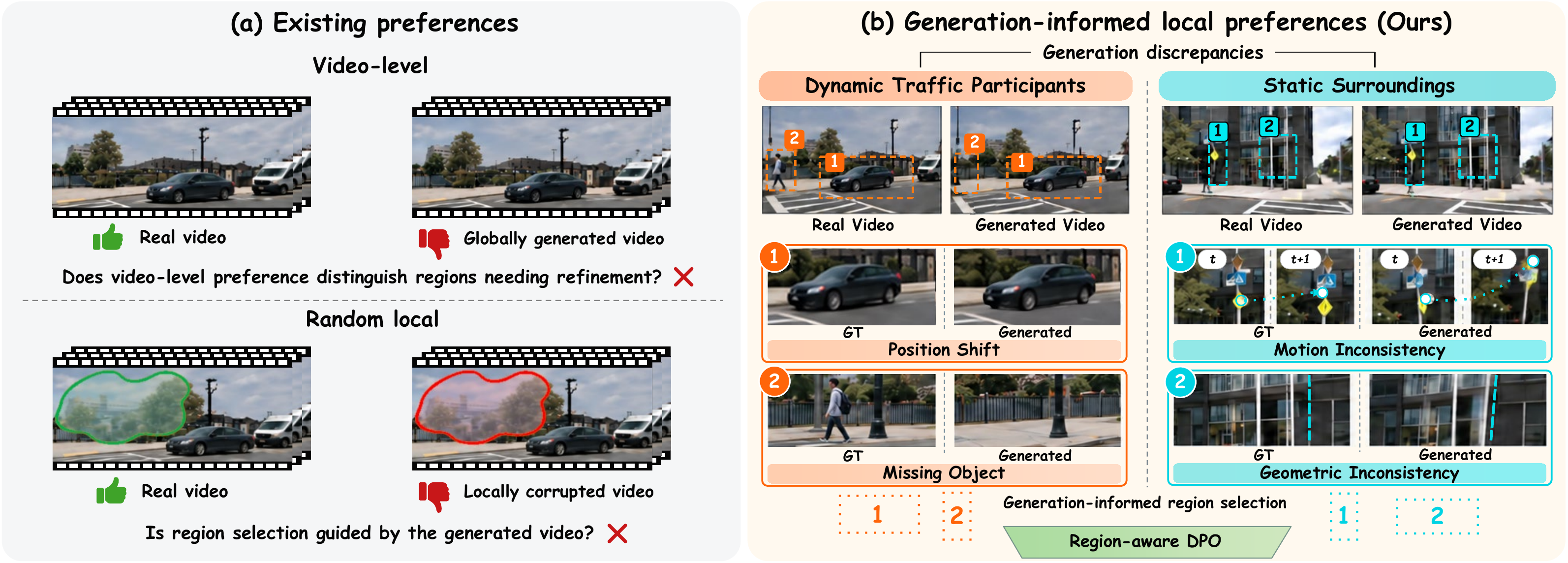}
        \captionof{figure}{\textbf{Comparison of preference construction strategies.}
    (a) Video-level preferences provide overall quality signals without explicitly distinguishing regions needing refinement, while random local methods select regions independently of generated video content.
    (b) GenLPO identifies object mismatches and cross-frame geometric inconsistencies to guide generation-informed region selection and verified local preference construction for region-aware DPO.}
        \label{fig:preference_overview}
    \end{minipage}
    \par\vspace{12pt}
}]
\begingroup
    \renewcommand{\thefootnote}{\fnsymbol{footnote}}
    \footnotetext[1]{Equal contributions.\quad
        \textsuperscript{\textdagger}Corresponding authors.}
\endgroup
\input{sec/0_abstract}    
\input{sec/1_intro}
\input{sec/2_related_work}
\input{sec/3_methodology}
\input{sec/4_experiments}
\input{sec/5_conclusion}

\section*{LLM Usage Statement}
We used OpenAI's ChatGPT to polish the language of the manuscript for grammar and clarity, and to assist in refining the related-work discussion, citations, code, and figures. The authors designed the method, conducted the experiments, analyzed the results, and verified the scientific content.

{
    \small
    \bibliographystyle{ieeenat_fullname}
    \bibliography{main}
}

\input{sec/appendix}

\end{document}

%% file: preamble.tex
\usepackage{multirow}
\usepackage{amssymb}
\usepackage{algorithm}
\usepackage{algpseudocode}
\definecolor{GenLPORowPink}{RGB}{252,230,250}
\definecolor{GenLPOGroupGray}{RGB}{235,235,235}
\definecolor{GenLPOHeaderGray}{RGB}{227,227,227}
\definecolor{GenLPORowBlue}{RGB}{238,250,252}
\definecolor{GenLPOChangePurple}{RGB}{112,38,108}

\usepackage{tabularx}
\usepackage{arydshln}
\IfPackageAtLeastT{array}{2026/06/01}{%
    \expandafter\let\csname endtabular*\endcsname\endtabular}
\usepackage{stfloats}
\DeclareMathSizes{6.5}{6}{5}{5}
\DeclareMathSizes{7.5}{7}{6}{5}

%% file: sec/0_abstract.tex
\begin{abstract}
Driving world models enable controllable synthesis of evolving traffic scenes for autonomous driving. While preference optimization offers a promising way to refine these models, video-level comparisons do not explicitly distinguish regions needing refinement, and random local selection ignores discrepancies in generated content.
In generated driving scenes, object mismatches involving dynamic traffic participants and cross-frame geometric inconsistencies in static surroundings provide complementary cues for targeted region selection.
Therefore, we propose \textbf{GenLPO}, a localized preference optimization framework that uses the model's own generations to guide both region selection and preference construction.
Specifically, we introduce \textbf{Object Mismatch Localization (OML)} to identify missing or displaced traffic participants through object-level spatial correspondence, and \textbf{Geometric Inconsistency Localization (GIL)} to localize static-scene discrepancies through cross-frame geometric analysis.
\textbf{Generation-Informed Preference Construction (GIPC)} constructs verified local preference pairs by replacing selected real-video regions with model-generated content while preserving surrounding context.
These preferences drive region-aware direct preference
optimization within an iterative generation-feedback loop.
Experiments on nuScenes demonstrate consistent improvements across multiple metrics on MagicDrive-V2 and UniMLVG, including FVD reductions of 17.5\% and 18.8\%, respectively.
Qualitative results further reveal clearer scene structures,
more complete traffic content, and improved object-level
visual fidelity.
\end{abstract}

%% file: sec/1_intro.tex
\section{Introduction}
\label{sec:intro}

Driving world models enable controllable synthesis of traffic scenarios for autonomous driving simulation and evaluation~\cite{10.1007/978-3-031-73195-2_4,Wang_2024_CVPR,ICLR2026_db6caae0,ICLR2026_a940e52c}. Recent advances in multi-view driving video generation~\cite{10.1007/978-3-031-73229-4_27,10.1007/978-3-031-72989-8_19,Ni_2025_CVPR,Wu_2025_CVPR}, including MagicDrive-V2~\cite{Gao_2025_ICCV} and UniMLVG~\cite{Chen_2025}, provide strong pretrained foundations for further refinement. Feedback-based post-training offers a complementary route through geometric rewards~\cite{NEURIPS2025_bb0f9af6} and preference optimization, ranging from video-level comparisons~\cite{Liu_2025_CVPR} and temporal segments~\cite{NEURIPS2025_fa975504} to localized preferences in LocalDPO~\cite{Huang_2026_CVPR}, constructed by regenerating randomly selected spatiotemporal regions of real videos. Together, these advances motivate more targeted preference optimization for driving world models.

Despite these advances, region selection and preference construction remain two distinct challenges in localized preference optimization. As illustrated in Fig.~\ref{fig:preference_overview}(a), video-level preferences~\cite{Liu_2025_CVPR} provide an overall quality signal but do not explicitly distinguish regions needing refinement. Random region selection enables local supervision~\cite{Huang_2026_CVPR}, but does not prioritize regions exhibiting discrepancies in generated videos. In driving scenes, these discrepancies include missing or displaced traffic participants and cross-frame geometric inconsistencies in static surroundings, as illustrated in Fig.~\ref{fig:preference_overview}(b). However, locating such discrepancies does not by itself yield preference pairs that represent the targeted generation failures. LocalDPO~\cite{Huang_2026_CVPR} constructs negative
samples by regenerating randomly masked regions of
real videos with a frozen video model, rather than
using content from regions exhibiting discrepancies
in the model's own generations.
This motivates constructing local preferences directly from the model's observed discrepancies, connecting region selection and preference construction through the same generation failures.

\begin{figure*}[t]
    \centering
    \includegraphics[width=\textwidth]{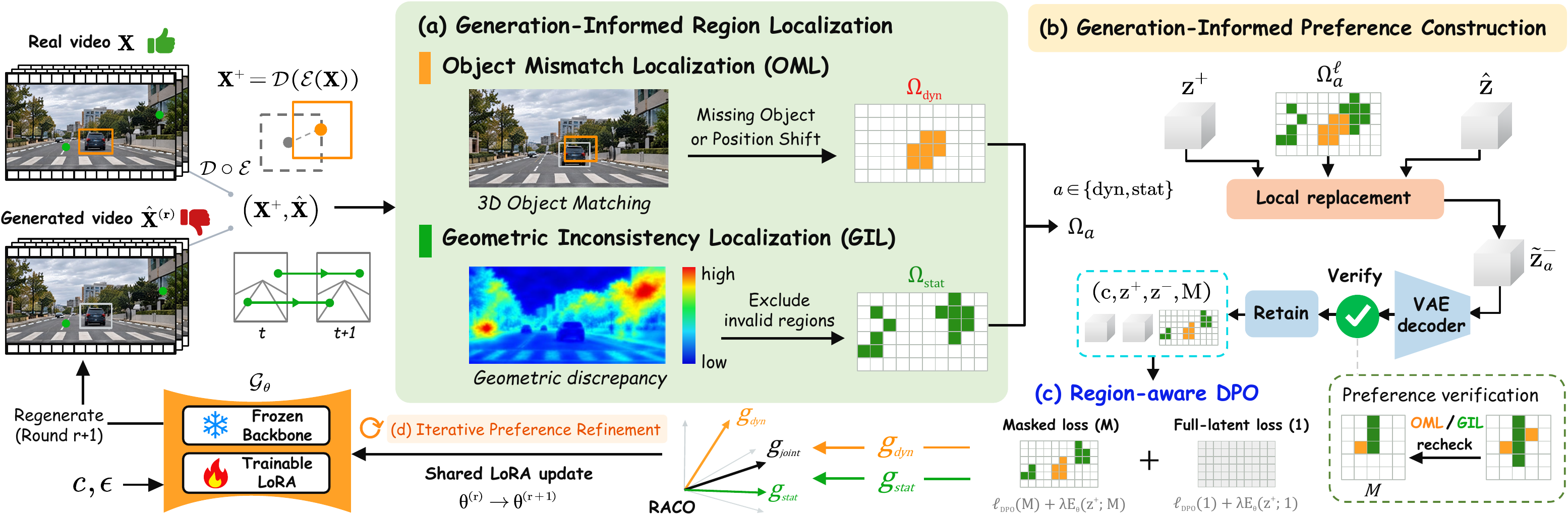}
    \caption{\textbf{Overview of GenLPO.}
    (a) Generation-Informed Region Localization.
    OML identifies missing or displaced traffic participants,
    while GIL locates cross-frame geometric inconsistencies.
    (b) Generation-Informed Preference Construction.
    Model-generated content is locally transplanted into preferred
    latents and verified after decoding.
    (c) Region-aware DPO.
    Masked and full-latent objectives are optimized with
    RACO-coordinated dynamic and static gradients.
    (d) Iterative Preference Refinement.
    Updated generations refresh region localization and
    verified preference pairs.}
    \label{fig:genlpo_framework}
\end{figure*}

To this end, we introduce \textbf{GenLPO}, a localized preference optimization framework for driving world models that uses the model's own generations to guide both region selection and preference construction.
As illustrated in Fig.~\ref{fig:genlpo_framework}, GenLPO integrates three complementary mechanisms for region selection and preference construction with region-aware optimization.
\textbf{Object Mismatch Localization (OML)} identifies missing or displaced traffic participants through spatial correspondence between conditioned 3D bounding boxes and detections in generated videos.
\textbf{Geometric Inconsistency Localization (GIL)} examines cross-frame geometric consistency in static surroundings under camera motion using optical flow, depth, and camera poses, selecting regions where generated videos are less consistent than real-video references.
\textbf{Generation-Informed Preference Construction (GIPC)}
constructs local negative samples by replacing selected
regions of ground-truth videos with corresponding
model-generated content, then verifies the samples and retains preference pairs only if the targeted
discrepancies persist.
Through an iterative generation-feedback loop, newly generated videos from the current model refresh region selection and local preference supervision, so that subsequent optimization targets discrepancies observed in its latest generations.

Experiments on nuScenes~\cite{Caesar_2020_CVPR} demonstrate improvements in generation quality and perception-based metrics on both MagicDrive-V2 and UniMLVG. Most notably, compared with their respective pretrained baselines, GenLPO reduces FVD~\cite{unterthiner2018towards} by \textbf{17.5\%} on MagicDrive-V2 and \textbf{18.8\%} on UniMLVG. Qualitative comparisons show that GenLPO suppresses spurious structures, recovers missing scene elements and traffic participants, and improves object-level visual fidelity, yielding cleaner, more complete, and more visually faithful driving videos.

Our main contributions are summarized as follows:
\begin{itemize}
\item We introduce \textbf{GenLPO}, a generation-informed localized preference optimization framework that leverages the model's own generations to adaptively refresh region selection and local preference supervision through an iterative generation-feedback loop.

\item We develop \textbf{OML} and \textbf{GIL} to localize object mismatches involving dynamic traffic participants and cross-frame geometric inconsistencies in static surroundings, enabling local supervision to focus on regions needing refinement.

\item We propose \textbf{GIPC} to construct verified local preference pairs from model-generated content, enabling region-aware DPO to learn directly from discrepancies observed in complete model-generated videos.

\item GenLPO achieves state-of-the-art generation quality among the compared driving world models on nuScenes, with FVD reductions of 17.5\% and 18.8\% on MagicDrive-V2 and UniMLVG, respectively, alongside improvements in perception-based metrics.

\end{itemize}

%% file: sec/2_related_work.tex
\section{Related work}
\label{sec:related_work}

\subsection{Driving world models}
\label{sec:driving_world_models}

Driving world models enable controllable traffic scene
generation for autonomous driving simulation and
data synthesis~\cite{hu2023gaia1generativeworldmodel,10.1007/978-3-031-73195-2_4,Wang_2024_CVPR}.
Early approaches use structured scene conditions
to guide controllable generation~\cite{10.1007/978-3-031-73195-2_4,ICLR2024_635351dc,Wen_2024_CVPR,10.1007/978-3-031-73229-4_27,10.1007/978-3-031-72989-8_19}.
Subsequent works advance spatially and temporally
consistent multi-view synthesis~\cite{Wang_2024_CVPR,mei2025dreamforgemotionawareautoregressivevideo,Jiang_2026,Ni_2025_CVPR,Wu_2025_CVPR,Dong_2025_ICCV}.
MagicDrive-V2~\cite{Gao_2025_ICCV}
combines spatiotemporal condition encoding
with multi-view diffusion transformers for
high-resolution videos,
while UniMLVG~\cite{Chen_2025}
integrates cross-view and temporal modeling
for controllable long-video generation.
Recent geometry-aware approaches incorporate
instance-level constraints, metric depth,
and 4D scene representations~\cite{Yang_2025_ICCV,Guo_2025_ICCV,Chen_2025_ICCV,Yang_2026_CVPR}.
Other approaches improve geometric consistency through geographic imagery~\cite{Jia_2026_CVPR} or geometry-aligned priors~\cite{10.1007/978-3-032-37531-5_22}, while RLGF~\cite{NEURIPS2025_bb0f9af6} uses geometric feedback for reinforcement learning-based post-training. These studies advance driving video generation through model design and geometric guidance, whereas our work focuses on refining pretrained generators through generation-informed local preference optimization.

\subsection{Preference learning for video generation}
\label{sec:preference_learning}

Preference-based optimization has become an important post-training approach to refining pretrained diffusion models. Building on direct preference optimization (DPO)~\cite{NEURIPS2023_a85b405e}, preference learning has been extended to diffusion and flow-matching generators~\cite{Wallace_2024_CVPR,Yang_2024_CVPR,li2026lineardpolineardirectpreference}, alongside reward-guided post-training for image and video generation~\cite{ICLR2024_14f75513,Yuan_2024_CVPR,prabhudesai2024videodiffusionalignmentreward,ICLR2025_e68af7d8}. For video generation, preference supervision draws on multi-dimensional quality assessments~\cite{Liu_2025_CVPR}, real-generated preference pairs~\cite{cheng2025realdporealrealpreference}, and online feedback~\cite{Zhang_2026_WACV}, while VideoAlign~\cite{NEURIPS2025_76227feb} explores multi-dimensional human feedback and flow-based optimization.

Recent studies investigate finer temporal
supervision through motion-aligned video pairs
and continuation-based comparisons~\cite{NEURIPS2025_fa975504,Li_2026_CVPR}, alongside region-level preferences~\cite{Huang_2026_CVPR,Lin_2026_CVPR}.
LocalDPO~\cite{Huang_2026_CVPR} constructs local
preferences by regenerating randomly selected
regions of real videos.
RiskMV-DPO~\cite{Lin_2026_CVPR} employs
motion-aware masks with 3D trajectory projections
to prioritize dynamic regions
for risk-controllable generation.
These studies refine preference supervision through temporal comparisons and region selection, with localized approaches using random or motion-guided masks. In contrast, our work selects regions from object mismatches and geometric inconsistencies observed in complete model-generated driving videos and constructs verified local preference pairs using the corresponding generated content for region-aware DPO.

%% file: sec/3_methodology.tex
\section{Methodology}
\label{sec:methodology}

\subsection{Overview}
\label{sec:overview}

Our method, \textbf{GenLPO}, refines pretrained driving world models through an iterative generation-feedback loop, as illustrated in Fig.~\ref{fig:genlpo_framework}. Let \(\mathbf{X}\) denote a real driving video and \(\mathbf{c}\) its scene conditions, including 3D bounding boxes, BEV maps, camera parameters, and other backbone-specific controls. At refinement round \(r\), the current generator with parameters \(\theta^{(r)}\) produces a video \(\hat{\mathbf{X}}^{(r)}\) from \(\mathbf{c}\) and sampling noise \(\boldsymbol{\epsilon}\). Omitting the round superscript for clarity, we write
\begin{equation}
\hat{\mathbf{X}}
=
\mathcal{G}_{\theta}
(\mathbf{c};\boldsymbol{\epsilon}),
\label{eq:driving_generation}
\end{equation}
where $\mathcal{G}_{\theta}$ denotes the complete video generation process of the selected pretrained generator. For each real video $\mathbf{X}$, we use the generator's VAE~\cite{ICLR2025_ce31378e,pmlr-v235-esser24a} to obtain the preferred latent $\mathbf{z}^{+}=\mathcal{E}(\mathbf{X})$ and its reconstruction $\mathbf{X}^{+}=\mathcal{D}(\mathbf{z}^{+})$, where $\mathcal{E}$ and $\mathcal{D}$ denote the encoder and decoder, respectively.
For region localization (Fig.~\ref{fig:genlpo_framework}(a), Sec.~\ref{sec:region_localization}), \textbf{OML} identifies missing or displaced traffic participants, while \textbf{GIL} identifies static regions where the generated video exhibits greater cross-frame geometric inconsistency than the preferred reconstruction. For preference construction (Fig.~\ref{fig:genlpo_framework}(b), Sec.~\ref{sec:local_preference_construction}), \textbf{GIPC} maps the selected spatiotemporal regions to the VAE latent grid and replaces the corresponding preferred latent content with content from the same model-generated video used for localization to construct candidate negative samples. We then decode each candidate and verify that the targeted discrepancies persist using the corresponding OML or GIL criteria. The verified regions define a supervision mask $\mathbf{M}$ for local preference optimization. Each resulting preference sample consists of $(\mathbf{c},\mathbf{z}^{+},\mathbf{z}^{-},\mathbf{M})$, where $\mathbf{z}^{-}$ denotes the rejected latent.

The verified local preferences guide region-aware DPO (Fig.~\ref{fig:genlpo_framework}(c)). After each optimization round, the updated generator produces new videos to refresh region localization and preference pairs (Fig.~\ref{fig:genlpo_framework}(d)), as detailed in Sec.~\ref{sec:iterative_preference_optimization}.

\subsection{Generation-Informed Region Localization}
\label{sec:region_localization}

\begin{figure}[t]
    \centering
    \includegraphics[width=\columnwidth]{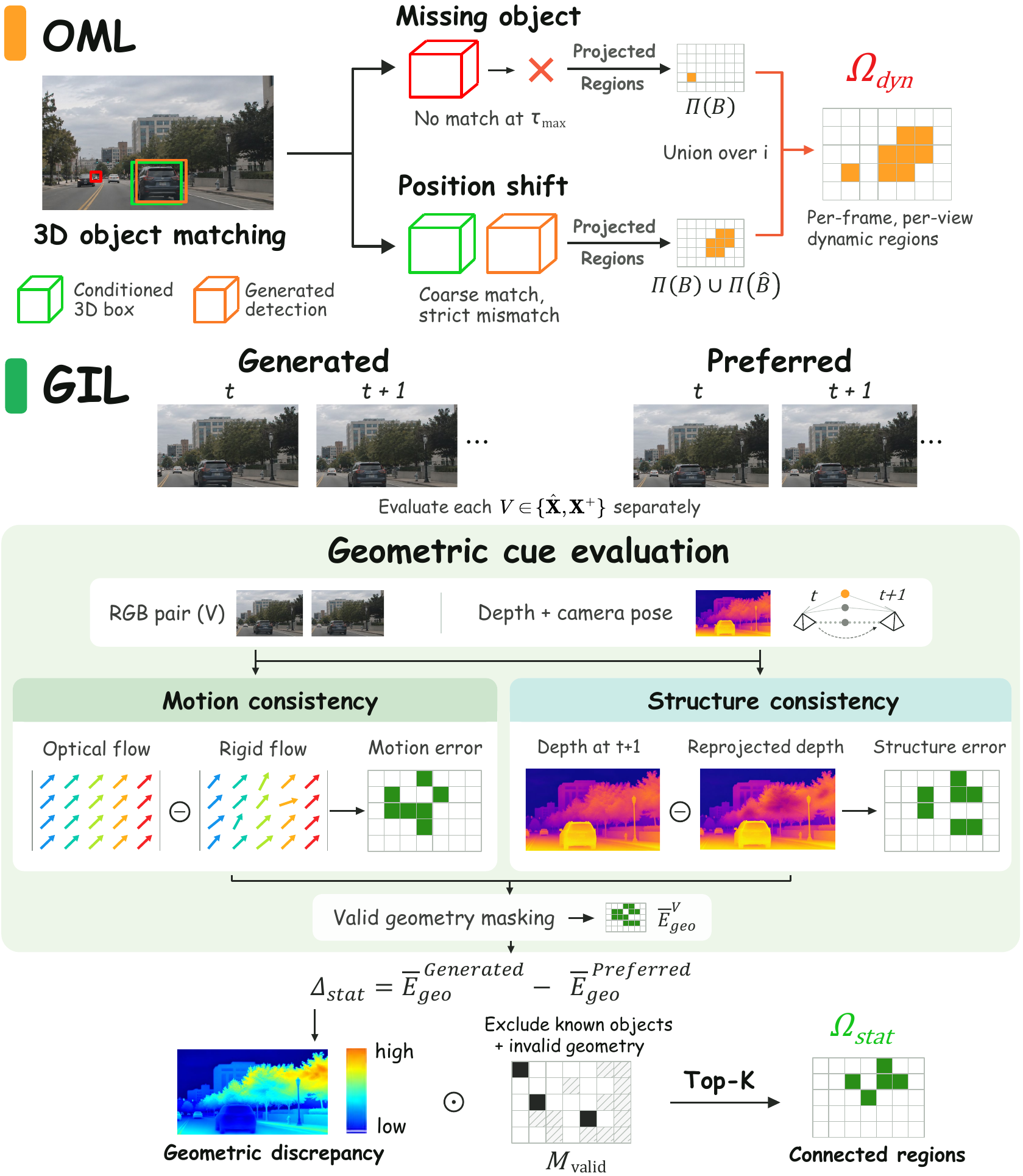}
    \caption{\textbf{Generation-informed region localization.}
    OML identifies missing or displaced traffic participants through 3D object matching.
    GIL selects static regions with greater cross-frame geometric inconsistency in generated videos than in preferred videos.}
    \label{fig:localization}
\end{figure}

As outlined in Fig.~\ref{fig:genlpo_framework}(a) and detailed in Fig.~\ref{fig:localization}, OML and GIL localize discrepancies in model-generated videos involving dynamic traffic participants and static surroundings, respectively. Both branches provide candidate regions for subsequent preference construction and verification.

\noindent\textbf{Object Mismatch Localization (OML).}
For dynamic traffic participants, we identify
discrepancies in object presence and position
through 3D object matching.
Specifically, we apply a pretrained
BEVFormer~\cite{10.1007/978-3-031-20077-9_1} 3D perception
model to the real video $\mathbf{X}$, the preferred
reconstruction $\mathbf{X}^{+}$, and the generated
video $\hat{\mathbf{X}}$.
Detected objects are matched to the conditioned
3D bounding boxes through class-consistent one-to-one
associations, using the center-distance thresholds
$\mathcal{T}$ of nuScenes~\cite{Caesar_2020_CVPR}.

For participant $i$ at frame $t$ and threshold $\tau\in\mathcal{T}$, let $r_{i,t}^{\tau}$, $p_{i,t}^{\tau}$, and $g_{i,t}^{\tau}$ be binary match indicators for $\mathbf{X}$, $\mathbf{X}^{+}$, and $\hat{\mathbf{X}}$, respectively ($1$ for a match and $0$ otherwise). We select participant-frame pairs satisfying
\begin{equation}
\begin{aligned}
r_{i,t}^{\tau}
&\le p_{i,t}^{\tau},
&&\forall\tau\in\mathcal{T},\\
(r_{i,t}^{\tau},p_{i,t}^{\tau},g_{i,t}^{\tau})
&=(1,1,0),
&&\exists\tau\in\mathcal{T}.
\end{aligned}
\label{eq:dynamic_candidates}
\end{equation}
These conditions require the preferred
reconstruction to preserve every real-video
match, while at least one such match is absent
from the generated video.

We distinguish \emph{Missing Object} and \emph{Position Shift} (Fig.~\ref{fig:localization}). Let $\mathbf{B}_{i,t}$ denote the conditioned 3D bounding box and $\hat{\mathbf{B}}_{i,t}$ the associated detection in the generated video, when available. \emph{Missing Object} refers to an expected participant with no corresponding detection near its conditioned location, even at the largest threshold $\tau_{\max}=\max\mathcal{T}$. \emph{Position Shift} refers to a participant whose generated detection is matched at $\tau_{\max}$ but remains unmatched under at least one smaller threshold in $\mathcal{T}$, indicating a displaced 3D center.

Their spatial regions in camera view $v$ are
defined as
\begin{equation}
\Omega_{i,t,v}^{\mathrm{dyn}}
=
\begin{cases}
\Pi_{t,v}(\mathbf{B}_{i,t}),
& \text{Missing Object},\\[3pt]
\Pi_{t,v}(\mathbf{B}_{i,t})
\cup
\Pi_{t,v}(\hat{\mathbf{B}}_{i,t}),
& \text{Position Shift},
\end{cases}
\label{eq:dynamic_regions}
\end{equation}
where $\Pi_{t,v}$ projects a 3D bounding box
onto camera view $v$ at frame $t$ and returns
its enclosing 2D rectangle.
For Missing Object, the conditioned box locates
the expected object region, whereas for Position Shift,
the union covers both the expected and generated
object locations.

We then take the union of selected participant regions at each frame and camera view to obtain the dynamic candidate regions $\{\Omega_{t,v}^{\mathrm{dyn}}\}_{t,v}$ for preference construction.

\noindent\textbf{Geometric Inconsistency Localization (GIL).} Unlike dynamic traffic participants, static surroundings such as roads, buildings, and roadside structures cannot be localized through the same instance-level 3D bounding box matching. However, the observed motion and depth relationships should remain consistent with the underlying static geometry, even when camera motion changes their image projections. We therefore use cross-frame geometric consistency to identify discrepancies in static regions.

Inspired by GeCo's motion and structure consistency cues~\cite{10.1007/978-3-032-37132-4_28}, we estimate depth and optical flow for $\hat{\mathbf{X}}$ and $\mathbf{X}^{+}$ using Depth Anything 3~\cite{ICLR2026_e4cd5012} and SEA-RAFT~\cite{10.1007/978-3-031-72667-5_3}, respectively. 
For each pixel $\mathbf{u}$ in frame $t$ and neighboring frame $s\in\mathcal{N}(t)$, where $\mathcal{N}(t)$ includes preceding and following frames, we compute two geometric consistency errors.

The \emph{motion consistency error} $E_{\mathrm{mot}}(t,s,\mathbf{u})$ measures the discrepancy between the estimated optical flow and the rigid flow computed from depth and relative camera pose, normalized by the focal length. The \emph{structural consistency error} $E_{\mathrm{str}}(t,s,\mathbf{u})$ measures the relative difference between the depth of the transformed 3D point and the estimated depth at its projected location in frame $s$. 
To identify static regions where the generated video is less geometrically consistent than the preferred reconstruction, we exclude known dynamic-object regions and unreliable correspondences, including invalid depth, inconsistent optical flow, and out-of-view projections. 
The detailed geometric projection and error calculations are provided in
Appendix~\ref{supp:geometric_consistency}.

We compute the motion and structural errors separately for $\hat{\mathbf{X}}$ and $\mathbf{X}^{+}$, combining only cues valid in both videos as
\begin{equation}
E_{\mathrm{geo}}(t,s,\mathbf{u})
=
\frac{1}{|\mathcal{C}_{t,s}(\mathbf{u})|}
\sum_{c\in\mathcal{C}_{t,s}(\mathbf{u})}
E_c(t,s,\mathbf{u}),
\label{eq:static_fused_error}
\end{equation}
where $\mathcal{C}_{t,s}(\mathbf{u})$ contains the jointly valid cues from $\{\mathrm{mot},\mathrm{str}\}$. We average both errors when available or use the sole valid error otherwise. Correspondences without jointly valid cues are discarded.

For each pixel $\mathbf{u}$ in frame $t$, we average the geometric errors over the same neighboring frames with at least one jointly valid cue in both videos, obtaining $\bar{E}_{\mathrm{geo}}^{\hat{\mathbf{X}}}(t,\mathbf{u})$ and $\bar{E}_{\mathrm{geo}}^{\mathbf{X}^{+}}(t,\mathbf{u})$. The pixelwise static discrepancy is defined as
\begin{equation}
\Delta_{\mathrm{stat}}(t,\mathbf{u})
=
\bar{E}_{\mathrm{geo}}^{\hat{\mathbf{X}}}(t,\mathbf{u})
-
\bar{E}_{\mathrm{geo}}^{\mathbf{X}^{+}}(t,\mathbf{u}).
\label{eq:static_discrepancy}
\end{equation}
A positive $\Delta_{\mathrm{stat}}$ indicates greater geometric inconsistency in the generated video at that pixel.

For latent-space preference construction, we aggregate positive pixelwise discrepancies into a heatmap aligned with the VAE latent grid.
For camera view $v$, $b$ indexes a temporal bin with frame set $\mathcal{I}_b$, and $\mathbf{q}$ indexes a spatial cell with pixel set $\mathcal{P}_{\mathbf{q}}$.
We average positive discrepancies
$\Delta_{\mathrm{stat}}^{(v)}$ over these frames
and pixels, with zero for positions lacking valid geometric
correspondences in either video:
\begin{equation}
H_{v,b}(\mathbf{q})
=
\frac{
\displaystyle\sum_{t\in\mathcal{I}_b}
\sum_{\mathbf{u}\in\mathcal{P}_{\mathbf{q}}}
[\Delta_{\mathrm{stat}}^{(v)}(t,\mathbf{u})]_+
}{
|\mathcal{I}_b|\,|\mathcal{P}_{\mathbf{q}}|
},
\label{eq:static_heatmap}
\end{equation}
where $[x]_+=\max(x,0)$.
Normalizing over all pixel-frame positions
captures both the magnitude and spatiotemporal
extent of geometric discrepancies.

As shown in Fig.~\ref{fig:localization}, for each camera $v$ and temporal bin $b$, we select the top-$k$ eligible cells as $\mathcal{Q}_{v,b}=\operatorname{TopK}_k(H_{v,b};\mathcal{A}_{v,b})$, where $\mathcal{A}_{v,b}$ contains cells with sufficient valid geometric evidence. We group adjacent selected cells into connected components and discard those with insufficient supporting observations.

The $j$-th retained component $\mathcal{K}_j\subseteq\mathcal{Q}_{v_j,b_j}$ defines a spatiotemporal region in camera $v_j$ and temporal bin $b_j$:
\begin{equation}
\Omega_j^{\mathrm{stat}}
=
\{v_j\}\times\mathcal{I}_{b_j}
\times
\bigcup_{\mathbf{q}\in\mathcal{K}_j}
\mathcal{P}_{\mathbf{q}}.
\label{eq:static_regions}
\end{equation}
The resulting $\{\Omega_j^{\mathrm{stat}}\}_j$ are the static candidate regions.

\subsection{Generation-Informed Preference Construction}
\label{sec:local_preference_construction}

OML and GIL locate object mismatches and geometric inconsistencies in complete model-generated videos. As shown in Fig.~\ref{fig:genlpo_framework}(b), GIPC constructs local negative samples using content from those same generated videos and verifies that the targeted discrepancies persist after decoding.

For branch $a\in\{\mathrm{dyn},\mathrm{stat}\}$, let $\Omega_a^{\ell}$ denote the selected OML or GIL regions mapped to the VAE latent grid.
Using the preferred latent $\mathbf{z}^{+}$
and the generated latent $\hat{\mathbf{z}}$
corresponding to the video $\hat{\mathbf{X}}$
used for localization, we construct a candidate
rejected latent through local replacement:
\begin{equation}
\begin{aligned}
\widetilde{\mathbf{z}}_a^{-}
&=
\mathbf{z}^{+}
+
\mathbf{1}_{\Omega_a^{\ell}}
\odot
(\hat{\mathbf{z}}-\mathbf{z}^{+}),\\
\widetilde{\mathbf{X}}_a^{-}
&=
\mathcal{D}(\widetilde{\mathbf{z}}_a^{-}),
\end{aligned}
\label{eq:local_preference_construction}
\end{equation}
where $\mathbf{1}_{\Omega_a^{\ell}}$
denotes the binary mask of the latent region,
and $\odot$ denotes element-wise multiplication.
The candidate incorporates model-generated content
within the selected region while preserving
the preferred latent outside it.

\noindent\textbf{Preference Verification.}
Region localization uses the complete generated video $\hat{\mathbf{X}}$, whereas preference optimization uses the constructed negative latent $\widetilde{\mathbf{z}}_a^{-}$. We therefore verify the targeted discrepancies again after local replacement and VAE decoding. For dynamic traffic participants, we reapply the 3D object-matching criterion in Eq.~\eqref{eq:dynamic_candidates} with $\widetilde{\mathbf{X}}_a^{-}$ replacing $\hat{\mathbf{X}}$. For static surroundings, we re-evaluate the cross-frame geometric discrepancy in Eq.~\eqref{eq:static_discrepancy} using $\widetilde{\mathbf{X}}_a^{-}$.

We retain only candidates passing this verification. For each retained candidate, we denote its rejected latent by $\mathbf{z}^{-}$ and use the verified regions to define the latent supervision mask $\mathbf{M}$. The resulting preference samples $(\mathbf{c},\mathbf{z}^{+},\mathbf{z}^{-},\mathbf{M})$ form the dataset $\mathcal{D}_{\mathrm{pref}}$. In this way, the retained preference pairs directly connect discrepancies observed in complete model generations with local preference supervision.

\begin{algorithm}[t]
\caption{Iterative Preference Optimization of GenLPO}
\label{alg:genlpo}
\begin{algorithmic}[1]
\Require Training data $\mathcal{D}_{\mathrm{real}}$;
initial policy $\theta^{(0)}$;
refinement rounds $R$;
update steps $\{K_r\}$
\Ensure Refined policy $\theta^{(R)}$

\For{$r=0,\ldots,R-1$}
    \State $\theta\gets\theta^{(r)}$
    \State $\theta_{\mathrm{ref}}
    \gets\theta^{(r)}$
    \Comment{Fixed within round}
    \State $\mathcal{D}_{\mathrm{pref}}^a
    \gets\varnothing$,
    $a\in\{\mathrm{dyn},\mathrm{stat}\}$

    \ForAll{$(\mathbf{X},\mathbf{c})
    \in\mathcal{D}_{\mathrm{real}}$}
        \State $\mathbf{z}^{+}
        \gets\mathcal{E}(\mathbf{X})$;
        $\mathbf{X}^{+}
        \gets\mathcal{D}(\mathbf{z}^{+})$

        \State Sample $\boldsymbol{\epsilon}$;
        $\hat{\mathbf{X}}
        \gets\mathcal{G}_{\theta}
        (\mathbf{c};\boldsymbol{\epsilon})$

        \State Retain $\hat{\mathbf{z}}$
        from the same generation

        \State $\Omega_{\mathrm{dyn}}
        \gets\operatorname{OML}
        (\mathbf{X},\mathbf{X}^{+},
        \hat{\mathbf{X}},\mathbf{c})$

        \State $\Omega_{\mathrm{stat}}
        \gets\operatorname{GIL}
        (\mathbf{X}^{+},
        \hat{\mathbf{X}},\mathbf{c})$

        \For{$a\in\{\mathrm{dyn},\mathrm{stat}\}$}
            \State Construct
            $\widetilde{\mathbf{z}}_a^{-}$
            via Eq.~\eqref{eq:local_preference_construction}

            \State Decode and verify
            $\widetilde{\mathbf{z}}_a^{-}$ via OML/GIL

            \State Store verified preference
            tuples in $\mathcal{D}_{\mathrm{pref}}^a$
        \EndFor
    \EndFor

    \For{$k=1,\ldots,K_r$}
        \For{$a\in\{\mathrm{dyn},\mathrm{stat}\}$}
            \State Sample minibatch
            $\mathcal{B}_a$
            from $\mathcal{D}_{\mathrm{pref}}^a$

            \State $\mathbf{g}_a
            \gets\nabla_{\theta}
            \frac{1}{|\mathcal{B}_a|}
            \sum_{d\in\mathcal{B}_a}
            \ell_{\theta}(d)$
        \EndFor

        \State $\mathbf{g}_{\mathrm{joint}}
        \gets\operatorname{RACO}
        (\mathbf{g}_{\mathrm{dyn}},
        \mathbf{g}_{\mathrm{stat}})$

        \State Update $\theta$
        using $\mathbf{g}_{\mathrm{joint}}$
    \EndFor

    \State $\theta^{(r+1)}\gets\theta$
\EndFor

\State \Return $\theta^{(R)}$
\end{algorithmic}
\end{algorithm}

\subsection{Iterative Preference Optimization}
\label{sec:iterative_preference_optimization}

At each refinement round, GenLPO regenerates
videos with the updated model, relocalizes
dynamic and static discrepancies, and rebuilds
verified preference pairs, as illustrated in
Fig.~\ref{fig:genlpo_framework}(d).
These preferences jointly guide the next round
of optimization, as summarized in
Algorithm~\ref{alg:genlpo}.

\noindent\textbf{Region-Aware Preference Optimization.}
For a verified preference sample
$d=(\mathbf{c},\mathbf{z}^{+},\mathbf{z}^{-},
\mathbf{M})$,
we adopt region-aware DPO~\cite{Wallace_2024_CVPR,Huang_2026_CVPR},
as illustrated in Fig.~\ref{fig:genlpo_framework}(c).
Let $f_{\theta}$ and $f_{\mathrm{ref}}$ denote
the current and frozen reference networks.
For a latent $\mathbf{z}$ at a sampled noise level $t$,
let $\mathbf{z}_t$ denote its noisy version
and $\mathbf{y}$ the model-specific prediction target.
The masked prediction error is
\begin{equation}
E_{\phi}(\mathbf{z};\mathbf{M})
=
\frac{
\sum_i M_i
\left(
f_{\phi,i}(\mathbf{z}_t,t,\mathbf{c})-y_i
\right)^2
}{
\sum_i M_i
},
\label{eq:masked_prediction_error}
\end{equation}
where $\phi\in\{\theta,\mathrm{ref}\}$,
$i$ indexes latent prediction elements,
and $\mathbf{M}$ is broadcast to the prediction shape.

Let
$\Delta_{\phi}(\mathbf{M})
=E_{\phi}(\mathbf{z}^{+};\mathbf{M})
-E_{\phi}(\mathbf{z}^{-};\mathbf{M})$.
The region-aware DPO loss is
\begin{equation}
\ell_{\mathrm{DPO}}(\mathbf{M})
=
\operatorname{softplus}
\left(
\beta
[
\Delta_{\theta}(\mathbf{M})
-
\Delta_{\mathrm{ref}}(\mathbf{M})
]
\right),
\label{eq:region_aware_dpo}
\end{equation}
where $\beta$ controls preference strength
and $\operatorname{softplus}(x)=\log(1+e^x)$.
We combine local and full-latent losses as
\begin{equation}
\begin{aligned}
\ell_{\theta}(d)
={}&
\ell_{\mathrm{DPO}}(\mathbf{M})
+
\ell_{\mathrm{DPO}}(\mathbf{1})
\\
&+
\lambda E_{\theta}(\mathbf{z}^{+};\mathbf{M})
+
\lambda E_{\theta}(\mathbf{z}^{+};\mathbf{1}),
\end{aligned}
\label{eq:structured_preference_loss}
\end{equation}
where $\mathbf{1}$ denotes the full latent region
and $\lambda$ weights the preferred-sample
prediction-error terms.

\input{sec/tables}

\noindent\textbf{Dynamic-Static Joint Optimization.}
We compute branch-specific gradients from dynamic and
static preferences using Eq.~\eqref{eq:structured_preference_loss}.
These gradients are combined using RACO~\cite{pmlr-v306-chen26bl}
to update a shared LoRA adapter~\cite{hu2022lora}.
Details of the gradient aggregation are provided in
Appendix~\ref{app:iterative_optimization}.

\GenLPODrivingSOTATable
\GenLPOPosttrainingTable

%% file: sec/tables.tex
\providecommand{\GenTBD}{\texttt{XX.XX}}
\providecommand{\GenNA}{\textemdash}
\providecommand{\GenLPOChange}[3][GenLPOChangePurple]{\textsubscript{\normalfont\color{#1}\ensuremath{#2}#3}}

\newcommand{\GenLPODrivingSOTATable}{%
\begin{table}[t]
\centering
\caption{
\textbf{Comparison with state-of-the-art driving world models on the nuScenes validation set.}
Results of prior methods are taken from
their original publications or subsequent
evaluations~\cite{Liang_2026_CVPR,Jiang_2026}.
Subscripts show relative performance changes (\%)
versus the best competing results.
}
\label{tab:driving_sota}

\small
\setlength{\tabcolsep}{1pt}
\renewcommand{\arraystretch}{1.08}
\newcommand{\GenLPOSOTAChange}[3][GenLPOChangePurple]{\textsubscript{\normalfont\fontsize{5}{6}\selectfont\color{##1}\ensuremath{##2}##3}}

\newcolumntype{Q}{>{\raggedright\arraybackslash}m{0.43\columnwidth}
                 *{4}{w{c}{\dimexpr(\columnwidth-0.43\columnwidth-10\tabcolsep)/4\relax}}}
\setlength{\fboxsep}{0pt}
\leavevmode\vbox{\offinterlineskip
\hbox{\colorbox{GenLPOHeaderGray}{%
\begin{tabular}{Q}
\specialrule{\heavyrulewidth}{\abovetopsep}{0pt}
\noalign{\vskip\belowrulesep}
\textbf{Method}
& \textbf{FID\boldmath$\downarrow$}
& \textbf{FVD\boldmath$\downarrow$}
& \textbf{mAP\boldmath$\uparrow$}
& \textbf{NDS\boldmath$\uparrow$} \\
\specialrule{\lightrulewidth}{\aboverulesep}{0pt}
\end{tabular}}}
\hbox{%
\begin{tabular}{Q}
\noalign{\vskip\belowrulesep}
Panacea~\cite{Wen_2024_CVPR}
& 16.96 & 139.00 & 11.58 & 22.31 \\
DreamForge~\cite{mei2025dreamforgemotionawareautoregressivevideo}
& 14.61 & 103.61
& 16.36 & 26.71 \\
MagicDrive-V2~\cite{Gao_2025_ICCV}
& 20.91 & 94.84 & 17.65 & 27.19 \\
NoiseController~\cite{Dong_2025_ICCV}
& 14.65 & 122.90 & 20.93 & 27.96 \\
UniMLVG~\cite{Chen_2025}
& 5.80 & 36.10 & \underline{22.50} & \underline{30.16} \\
GeoFlow~\cite{10.1007/978-3-032-37531-5_22}
& 6.80 & 32.50 & \GenNA & \GenNA \\
RiskMV-DPO~\cite{Lin_2026_CVPR}
& 15.70 & 87.65 & \GenNA & \GenNA \\
MagicDrive-V2 +~Geo~\cite{Jia_2026_CVPR}
& 18.10 & 81.52 & \GenNA & \GenNA \\
UniMLVG +~Geo~\cite{Jia_2026_CVPR}
& \underline{5.60} & \underline{29.97} & \GenNA & \GenNA \\
\midrule
\rowcolor{GenLPORowPink}
\textbf{MagicDrive-V2 + Ours}
& 14.54 & 78.24
& 18.39 & 28.37 \\
\rowcolor{GenLPORowPink}
\textbf{UniMLVG + Ours}
& \textbf{5.23}\GenLPOSOTAChange{\uparrow}{6.61}
& \textbf{29.33}\GenLPOSOTAChange{\uparrow}{2.14}
& \textbf{23.17}\GenLPOSOTAChange{\uparrow}{2.98}
& \textbf{31.09}\GenLPOSOTAChange{\uparrow}{3.08} \\
\bottomrule
\end{tabular}}
}
\end{table}
}

\newcommand{\GenLPOPosttrainingTable}{%
\begin{table*}[t]
\centering
\caption{
\textbf{Comparison with different post-training methods
on MagicDrive-V2 and UniMLVG.}
All methods within each backbone are evaluated
under identical conditions.
In the CVT group, $\mathrm{mIoU}_{r}$ and
$\mathrm{mIoU}_{v}$ denote road and vehicle mIoU, respectively.
Subscripts show relative improvements (\%) over the baseline
within each backbone.
}
\label{tab:posttraining}
\normalsize
\setlength{\tabcolsep}{3pt}
\renewcommand{\arraystretch}{1.12}
\def\GenLPOVLineTopGap{0pt}
\def\GenLPOVLineBottomGap{0pt}
\newcommand{\GenLPOGroupVLine}{\kern-\arrayrulewidth{\color{black}%
  \vrule width\arrayrulewidth
  height\dimexpr\arraystretch\ht\strutbox+\arraystretch\extrarowheight-\GenLPOVLineTopGap\relax
  depth\dimexpr\arraystretch\dp\strutbox-\GenLPOVLineBottomGap\relax}}

\newcolumntype{J}{w{l}{0.15\textwidth}!{\GenLPOGroupVLine}
  *{3}{*{2}{w{c}{\dimexpr(\textwidth-0.15\textwidth-18\tabcolsep)/8\relax}}!{\GenLPOGroupVLine}}
  *{2}{w{c}{\dimexpr(\textwidth-0.15\textwidth-18\tabcolsep)/8\relax}}}
\setlength{\fboxsep}{0pt}
\leavevmode\vbox{\offinterlineskip
\hbox{\colorbox{GenLPOHeaderGray}{%
\begin{tabular}{J}
\specialrule{\heavyrulewidth}{\abovetopsep}{0pt}
\noalign{\vskip\belowrulesep}

& \multicolumn{2}{c!{\GenLPOGroupVLine}}{\textbf{Quality}}
& \multicolumn{2}{c!{\GenLPOGroupVLine}}{\textbf{BEVFormer}}
& \multicolumn{2}{c!{\GenLPOGroupVLine}}{\textbf{BEVFusion}}
& \multicolumn{2}{c}{\textbf{CVT}} \\
\cmidrule(lr){2-3}\cmidrule(lr){4-5}\cmidrule(lr){6-7}\cmidrule(lr){8-9}
\multicolumn{1}{w{c}{0.15\textwidth}!{\GenLPOGroupVLine}}{%
  \multirow{-2}{*}[\dimexpr-0.8ex+\aboverulesep+\cmidrulewidth+\belowrulesep\relax]{\textbf{Method}}}
& \textbf{FID\boldmath$\downarrow$}
& \textbf{FVD\boldmath$\downarrow$}
& \textbf{mAP\boldmath$\uparrow$}
& \textbf{mIoU\boldmath$\uparrow$}
& \textbf{mAP\boldmath$\uparrow$}
& \textbf{NDS\boldmath$\uparrow$}
& \textbf{\boldmath$\mathrm{mIoU}_{r}\uparrow$}
& \textbf{\boldmath$\mathrm{mIoU}_{v}\uparrow$} \\

\specialrule{\lightrulewidth}{\aboverulesep}{0pt}
\end{tabular}}}
\hbox{%
\begin{tabular}{J}
\noalign{\vskip\belowrulesep}
\rowcolor{GenLPOGroupGray}
\multicolumn{9}{c}{\textit{MagicDrive-V2}} \\

\noalign{\gdef\GenLPOVLineTopGap{1.2pt}}
Baseline
& 20.91 & 94.84
& 18.17 & 20.40
& 17.65 & 27.19 & 59.79 & 32.73 \\
\noalign{\gdef\GenLPOVLineTopGap{0pt}}

SFT
& 19.21 & 92.75
& 18.06 & 20.53
& 17.58 & 27.51 & 59.67 & 31.97 \\

Vanilla DPO~\cite{Liu_2025_CVPR}
& 18.98 & 91.13
& 18.10 & 20.71
& \underline{17.72} & \underline{28.03} & 59.88 & 32.65 \\

LocalDPO~\cite{Huang_2026_CVPR}
& \underline{17.53} & \underline{87.09}
& \underline{18.21} & \underline{20.86}
& 17.62 & 27.89 & \underline{60.13} & \underline{32.74} \\

\noalign{\gdef\GenLPOVLineBottomGap{1.2pt}}
\rowcolor{GenLPORowPink}
\textbf{GenLPO (Ours)}
& \textbf{14.54}\GenLPOChange{\uparrow}{30.46} & \textbf{78.24}\GenLPOChange{\uparrow}{17.50}
& \textbf{18.75}\GenLPOChange{\uparrow}{3.19} & \textbf{21.92}\GenLPOChange{\uparrow}{7.45}
& \textbf{18.39}\GenLPOChange{\uparrow}{4.19} & \textbf{28.37}\GenLPOChange{\uparrow}{4.34} & \textbf{60.95}\GenLPOChange{\uparrow}{1.94} & \textbf{33.26}\GenLPOChange{\uparrow}{1.62} \\
\noalign{\gdef\GenLPOVLineBottomGap{0pt}}

\midrule
\rowcolor{GenLPOGroupGray}
\multicolumn{9}{c}{\textit{UniMLVG}} \\

\noalign{\gdef\GenLPOVLineTopGap{1.2pt}}
Baseline
& 5.80 & 36.10
& 24.03 & 35.12
& 22.50 & 30.16 & 70.81 & 29.12 \\
\noalign{\gdef\GenLPOVLineTopGap{0pt}}

SFT
& 5.71 & 35.19
& 23.78 & 35.15 & 22.27 & 30.24 & \underline{70.96} & 29.23 \\

Vanilla DPO~\cite{Liu_2025_CVPR}
& 5.68 & 35.04
& \underline{24.19} & 35.08 & 22.46 & 30.28 & 70.59 & 29.40 \\

LocalDPO~\cite{Huang_2026_CVPR}
& \underline{5.54} & \underline{34.12}
& 24.11 & \underline{35.21} & \underline{22.61} & \underline{30.35} & 70.64 & \underline{29.51} \\

\noalign{\gdef\GenLPOVLineBottomGap{1.2pt}}
\rowcolor{GenLPORowPink}
\textbf{GenLPO (Ours)}
& \textbf{5.23}\GenLPOChange{\uparrow}{9.83} & \textbf{29.33}\GenLPOChange{\uparrow}{18.75}
& \textbf{24.95}\GenLPOChange{\uparrow}{3.83} & \textbf{35.74}\GenLPOChange{\uparrow}{1.77}
& \textbf{23.17}\GenLPOChange{\uparrow}{2.98} & \textbf{31.09}\GenLPOChange{\uparrow}{3.08} & \textbf{71.10}\GenLPOChange{\uparrow}{0.41} & \textbf{31.16}\GenLPOChange{\uparrow}{7.01} \\
\noalign{\gdef\GenLPOVLineBottomGap{0pt}}

\bottomrule
\end{tabular}}
}
\end{table*}
}

\newcommand{\GenLPOAblationTable}{%
\begin{table}[t]
\centering
\caption{\textbf{Progressive ablation of GenLPO on MagicDrive-V2.}
Iter. denotes iterative preference refinement. mAP and NDS are evaluated using BEVFusion.}
\label{tab:ablation}
\small
\setlength{\tabcolsep}{1pt}
\renewcommand{\arraystretch}{1.12}
\newcolumntype{R}{*{4}{w{c}{\dimexpr0.10\columnwidth-2\tabcolsep\relax}}
                 *{4}{w{c}{\dimexpr0.15\columnwidth-2\tabcolsep\relax}}}
\setlength{\fboxsep}{0pt}
\leavevmode\vbox{\offinterlineskip
\hbox{\colorbox{GenLPOHeaderGray}{%
\begin{tabular}{R}
\specialrule{\heavyrulewidth}{\abovetopsep}{0pt}
\noalign{\vskip\belowrulesep}
\textbf{OML} & \textbf{GIL} & \textbf{GIPC} & \textbf{Iter.}
& \textbf{FID\boldmath$\downarrow$} & \textbf{FVD\boldmath$\downarrow$} & \textbf{mAP\boldmath$\uparrow$} & \textbf{NDS\boldmath$\uparrow$} \\
\specialrule{\lightrulewidth}{\aboverulesep}{0pt}
\end{tabular}}}
\hbox{%
\begin{tabular}{R}
\noalign{\vskip\belowrulesep}
& & &
& 20.91 & 94.84 & 17.65 & 27.19 \\
$\checkmark$ & & &
& 17.92 & 87.71 & 17.96 & 27.81 \\
$\checkmark$ & $\checkmark$ & &
& 16.35 & 83.93 & 18.01 & 27.85 \\
$\checkmark$ & $\checkmark$ & $\checkmark$ &
& \underline{15.77} & \underline{81.36} & \underline{18.22} & \underline{28.10} \\
\rowcolor{GenLPORowPink}
$\checkmark$ & $\checkmark$ & $\checkmark$ & $\checkmark$
& \textbf{14.54} & \textbf{78.24} & \textbf{18.39} & \textbf{28.37} \\
\bottomrule
\end{tabular}}
}
\end{table}
}


\definecolor{suppresultheader}{HTML}{E3E3E3}
\definecolor{suppresultrow}{HTML}{FCE6FA}
\definecolor{suppresultaccent}{HTML}{70266C}
\newcommand{\suppresultdivider}{%
    \kern-\arrayrulewidth{\color{black!65}\vrule width\arrayrulewidth}}

\newcommand{\GenLPOSuppNotationTable}{%
\begin{table}[t]
\centering
\caption{Notation used in this supplement.}
\label{supp:tab:notation}
\small
\setlength{\tabcolsep}{3.5pt}
\renewcommand{\arraystretch}{1.12}
\setlength{\dashlinedash}{3pt}
\setlength{\dashlinegap}{3pt}
\definecolor{notationheader}{HTML}{E3E3E3}
\newcommand{\notationgroup}[1]{%
    \multicolumn{2}{l}{\color{black!60}\itshape ##1}\\}
\newcommand{\notationstripe}{\rowcolor{black!5}}
\newcommand{\notationheaderheight}{16pt}
\begin{tabularx}{\columnwidth}{>{\centering\arraybackslash}p{0.27\columnwidth}
                              >{\raggedright\arraybackslash}X}
\specialrule{\heavyrulewidth}{0pt}{0pt}
\noalign{%
    {\color{notationheader}\hrule height\notationheaderheight}%
    \vskip-\notationheaderheight}
\rule[-5pt]{0pt}{\notationheaderheight}\textbf{Notation} & \textbf{Description} \\
\specialrule{\lightrulewidth}{0pt}{\belowrulesep}
\notationgroup{Driving Videos}
\notationstripe $\mathbf{X}$ & Real driving video \\
$\mathbf{X}^{+}$ & Preferred VAE reconstruction \\
\notationstripe $\hat{\mathbf{X}}$ & Complete model-generated video \\
\hdashline
\notationgroup{Object Matching and Projection}
\notationstripe $\mathcal{T}$ & BEV center-distance thresholds \\
$r_{i,t}^{\tau}$ & Real-video match indicator \\
\notationstripe $p_{i,t}^{\tau}$ & Preferred-video match indicator \\
$g_{i,t}^{\tau}$ & Generated-video match indicator \\
\notationstripe $\mathbf{B}_{i,t},\hat{\mathbf{B}}_{i,t}$ & Conditioned and detected 3D boxes \\
$\Pi_{t,v}$ & 3D-box projection to a 2D rectangle \\
\notationstripe $\Omega_{t,v}^{\mathrm{dyn}}$ & Dynamic candidate region \\
\hdashline
\notationgroup{Geometric Consistency}
\notationstripe $D_t(\mathbf{u})$ & Estimated depth at pixel $\mathbf{u}$ \\
$\mathbf{F}_{t\to s}(\mathbf{u})$ & Optical flow from frame $t$ to $s$ \\
\notationstripe $\mathbf{p}_s(\mathbf{u})$ & 3D point in target camera frame \\
$\mathbf{u}'$ & Rigidly reprojected pixel \\
\notationstripe $E_{\mathrm{mot}},E_{\mathrm{str}}$ & Motion and structure errors \\
$\mathcal{C}_{t,s}(\mathbf{u})$ & Cues valid in both videos \\
\notationstripe $H_{v,b}(\mathbf{q})$ & Nonnegative latent-cell heatmap score \\
\hdashline
\notationgroup{Preference Optimization}
\notationstripe $\mathbf{M}$ & Supervision mask over verified regions \\
$\beta$ & DPO scaling coefficient \\
\notationstripe $\lambda$ & Preferred-sample flow-matching weight \\
\hdashline
\notationgroup{Gradient Aggregation}
\notationstripe $g_{\mathrm{dyn}},g_{\mathrm{stat}}$ & Dynamic and static branch gradients \\
$g_0,g(p)$ & Mean and weighted branch gradients \\
\notationstripe $p^*$ & Optimized mixing coefficient \\
$g_{\mathrm{joint}}$ & Aggregated gradient for shared LoRA \\
\notationstripe $\gamma$ & RACO aggregation coefficient \\
\bottomrule
\end{tabularx}
\end{table}%
}

\newcommand{\GenLPOSuppOMLMatchingTable}{%
\begin{table}[t]
\centering
\captionsetup{skip=7pt}
\caption{\textbf{OML Matching Cases Across Thresholds.} The five cases use constructed BEV center distances between the conditioned box and same-class detections. A dash indicates no detection candidate. Each triplet denotes $(r_{i,t}^{\tau},p_{i,t}^{\tau},g_{i,t}^{\tau})$.}
\label{supp:tab:oml_matching}
\fontsize{6.5pt}{9pt}\selectfont
\renewcommand{\arraystretch}{1.10}
\setlength{\tabcolsep}{2pt}
\begin{tabular}{
  w{c}{\dimexpr0.092\columnwidth-2\tabcolsep\relax}!{\suppresultdivider}
  *{3}{w{c}{\dimexpr0.075\columnwidth-2\tabcolsep\relax}}!{\suppresultdivider}
  *{4}{w{c}{\dimexpr0.122\columnwidth-2\tabcolsep\relax}}!{\suppresultdivider}
  >{\hspace{1.25pt}}w{l}{\dimexpr0.195\columnwidth-2\tabcolsep\relax}}
\toprule
& \multicolumn{3}{c!{\suppresultdivider}}{BEV dist. (m)}
& \multicolumn{4}{c!{\suppresultdivider}}{Match at $\tau$ (m)}
& \\
\cmidrule(lr){2-4}
\cmidrule(lr){5-8}
\multirow{-2}{*}[\dimexpr(\aboverulesep+\cmidrulewidth+\belowrulesep)/2\relax]{\textbf{Case}}
     & $\mathbf{X}$ & $\mathbf{X}^{+}$ & $\hat{\mathbf{X}}$
     & $0.5$ & $1$ & $2$ & $4$
     & \multicolumn{1}{c}{%
         \multirow{-2}{*}[\dimexpr(\aboverulesep+\cmidrulewidth+\belowrulesep)/2\relax]{\textbf{Outcome}}} \\
\midrule
1 & $0.3$ & $0.4$ & --
  & $(1,1,0)$ & $(1,1,0)$ & $(1,1,0)$ & $(1,1,0)$ & Missing Object \\
2 & $0.3$ & $0.4$ & $1.6$
  & $(1,1,0)$ & $(1,1,0)$ & $(1,1,1)$ & $(1,1,1)$ & Position Shift \\
3 & $0.7$ & $0.3$ & $1.6$
  & $(0,1,0)$ & $(1,1,0)$ & $(1,1,1)$ & $(1,1,1)$ & Position Shift \\
4 & $0.3$ & $0.7$ & $1.6$
  & $(1,0,0)$ & $(1,1,0)$ & $(1,1,1)$ & $(1,1,1)$ & Not selected \\
5 & $0.3$ & $0.4$ & $0.2$
  & $(1,1,1)$ & $(1,1,1)$ & $(1,1,1)$ & $(1,1,1)$ & Not selected \\
\bottomrule
\end{tabular}
\end{table}%
}

\newcommand{\GenLPOSuppRoundwiseTable}{%
\begin{table}[!t]
\centering
\captionsetup{skip=7pt}
\caption{\textbf{Generation quality across rounds on nuScenes.}
Rounds 1--4 end at 200, 300, 400, and 500 cumulative steps.
\mbox{Arrows} show relative improvements (\%) over Round~1.}
\label{supp:tab:roundwise_optimization}
\small
\setlength{\tabcolsep}{3.5pt}
\renewcommand{\arraystretch}{1.12}
\newcommand{\roundtableimprovement}[1]{%
    \textsubscript{\normalfont\color{suppresultaccent}\ensuremath{\uparrow}##1}}
\newcolumntype{R}{w{c}{\dimexpr0.20\columnwidth-2\tabcolsep\relax}!{\suppresultdivider}
                 *{2}{w{c}{\dimexpr0.20\columnwidth-2\tabcolsep\relax}}!{\suppresultdivider}
                 *{2}{w{c}{\dimexpr0.20\columnwidth-2\tabcolsep\relax}}}
\setlength{\fboxsep}{0pt}
\leavevmode\vbox{\offinterlineskip
\hbox{\colorbox{suppresultheader}{%
\begin{tabular}{R}
\specialrule{\heavyrulewidth}{0pt}{0pt}
\noalign{\vskip\belowrulesep}
& \multicolumn{2}{c!{\suppresultdivider}}{\textbf{MagicDrive-V2}}
& \multicolumn{2}{c}{\textbf{UniMLVG}} \\
\cmidrule(lr){2-3}\cmidrule(lr){4-5}
\multirow{-2}{*}[\dimexpr(\aboverulesep+\cmidrulewidth+\belowrulesep)/2\relax]{\textbf{Stage}}
& \textbf{FID\boldmath$\downarrow$} & \textbf{FVD\boldmath$\downarrow$}
& \textbf{FID\boldmath$\downarrow$} & \textbf{FVD\boldmath$\downarrow$} \\
\specialrule{\lightrulewidth}{\aboverulesep}{0pt}
\end{tabular}}}
\hbox{%
\begin{tabular}{R}
\noalign{\vskip\belowrulesep}
Pretrained & 20.91 & 94.84 & 5.80 & 36.10 \\
Round~1 & 16.05 & 82.60 & 5.49 & 30.92 \\
Round~2 & 15.42 & 80.53 & 5.34 & 30.75 \\
Round~3 & 14.80 & 78.79 & 5.31 & 30.04 \\
\rowcolor{suppresultrow}
\textbf{Round~4} & \textbf{14.54}\roundtableimprovement{9.41}
                 & \textbf{78.24}\roundtableimprovement{5.28}
                 & \textbf{5.23}\roundtableimprovement{4.74}
                 & \textbf{29.33}\roundtableimprovement{5.14} \\
\specialrule{\heavyrulewidth}{\belowrulesep}{0pt}
\end{tabular}}
}
\end{table}%
}

\makeatletter
\newcommand{\GenLPOSuppAdditionalMetricsTable}{%
\if@firstcolumn
    \suppressfloats[t]%
\fi
\begin{table}[t]
\centering
\captionsetup{skip=7pt}
\caption{\textbf{Object-level accuracy and geometric consistency.}
\mbox{Arrows} show relative improvements (\%) over each pretrained model.}
\label{supp:tab:additional_metrics}
\fontsize{7.5pt}{9pt}\selectfont
\setlength{\tabcolsep}{1.25pt}
\renewcommand{\arraystretch}{1.20}
\newcommand{\metricsimprovement}[1]{%
    \textsubscript{\normalfont\tiny\color{suppresultaccent}\ensuremath{\uparrow}##1}}
\newcolumntype{Q}{w{l}{\dimexpr0.33\columnwidth-2\tabcolsep\relax}!{\suppresultdivider}
                 *{2}{w{c}{\dimexpr0.1675\columnwidth-2\tabcolsep\relax}}!{\suppresultdivider}
                 *{2}{w{c}{\dimexpr0.1675\columnwidth-2\tabcolsep\relax}}}
\setlength{\fboxsep}{0pt}
\leavevmode\vbox{\offinterlineskip
\hbox{\colorbox{suppresultheader}{%
\begin{tabular}{Q}
\specialrule{\heavyrulewidth}{0pt}{0pt}
\noalign{\vskip\belowrulesep}
& \multicolumn{2}{c!{\suppresultdivider}}{\textbf{Object Accuracy}}
& \multicolumn{2}{c}{\textbf{Static Geometry}} \\
\cmidrule(lr){2-3}\cmidrule(lr){4-5}
\multirow{-2}{*}[\dimexpr(\aboverulesep+\cmidrulewidth+\belowrulesep)/2\relax]{%
    \makebox[\dimexpr0.33\columnwidth-2\tabcolsep\relax][c]{\textbf{Method}}}
& \textbf{AP@1m}$\uparrow$ & \textbf{mATE}$\downarrow$
& \textbf{Motion}$\downarrow$ & \textbf{Structure}$\downarrow$ \\
\specialrule{\lightrulewidth}{\aboverulesep}{0pt}
\end{tabular}}}
\hbox{%
\begin{tabular}{Q}
\noalign{\vskip\belowrulesep}
MagicDrive-V2~\cite{Gao_2025_ICCV} & 11.85\% & 0.853 & 0.0160 & 0.0416 \\
\rowcolor{suppresultrow}
\textbf{MagicDrive-V2 + Ours} & \textbf{12.49\%}\metricsimprovement{5.40}
& \textbf{0.837}\metricsimprovement{1.88}
& \textbf{0.0143}\metricsimprovement{10.63}
& \textbf{0.0388}\metricsimprovement{6.73} \\
\midrule
UniMLVG~\cite{Chen_2025} & 11.34\% & 0.866 & 0.0119 & 0.0407 \\
\rowcolor{suppresultrow}
\textbf{UniMLVG + Ours} & \textbf{11.95\%}\metricsimprovement{5.38}
& \textbf{0.845}\metricsimprovement{2.42}
& \textbf{0.0107}\metricsimprovement{10.08}
& \textbf{0.0383}\metricsimprovement{5.90} \\
\specialrule{\heavyrulewidth}{\belowrulesep}{0pt}
\end{tabular}}
}
\end{table}%
}
\makeatother

%% file: sec/4_experiments.tex
\section{Experiments}
\label{sec:experiments}

\subsection{Experimental Setup}

\noindent\textbf{Dataset and Backbones.}
We conduct experiments on the nuScenes dataset~\cite{Caesar_2020_CVPR},
following the official training and validation splits of 700 and 150
driving scenes, respectively.
The dataset provides six surround-view camera streams,
3D object annotations, HD maps, and camera calibration parameters,
which are used to construct the scene conditions for controllable generation.
We evaluate GenLPO on two representative pretrained driving world models, MagicDrive-V2~\cite{Gao_2025_ICCV}
and UniMLVG~\cite{Chen_2025}.
For both backbones, we construct local preference pairs using
the training split and evaluate the refined models on the
held-out validation split.
We preserve the original backbone architectures and
their respective conditioning pipelines.

\noindent\textbf{Evaluation Metrics.}
We evaluate generation quality using
Fr\'echet Video Distance (FVD)~\cite{unterthiner2018towards}
and Fr\'echet Inception Distance (FID)~\cite{NIPS2017_8a1d6947},
which measure distributional differences in video and image
feature spaces, respectively.
To assess controllability, we use
BEVFormer~\cite{10.1007/978-3-031-20077-9_1}
for 3D object detection and bird's-eye-view (BEV) map segmentation,
BEVFusion~\cite{Liu_2023} for detection,
and Cross-View Transformers (CVT)~\cite{Zhou_2022_CVPR}
for road and vehicle segmentation.
We report mean average precision (mAP) for detection,
mean intersection-over-union (mIoU) for segmentation,
and nuScenes detection score (NDS)~\cite{Caesar_2020_CVPR}
computed with BEVFusion, all as percentages.
For each backbone, pretrained and fine-tuned models share
the same validation data, generation settings, metric
implementations, and perception evaluators, following its
original protocol~\cite{Gao_2025_ICCV,Chen_2025}.

\noindent\textbf{Implementation Details.}
We perform parameter-efficient post-training on
MagicDrive-V2 and UniMLVG using LoRA~\cite{hu2022lora}
with rank 32 and $\alpha=64$ at a learning rate of
$8\times10^{-6}$, while keeping their pretrained
backbones and VAEs frozen.
For region localization, OML employs a pretrained BEVFormer~\cite{10.1007/978-3-031-20077-9_1} model, while GIL uses Depth Anything 3~\cite{ICLR2026_e4cd5012} and SEA-RAFT~\cite{10.1007/978-3-031-72667-5_3} for depth and optical-flow estimation, respectively.
All perception models remain frozen, and localized preference pairs are verified after VAE decoding.
All experiments are conducted on NVIDIA A100 GPUs.
Further details are provided in Appendix~\ref{supp:experimental_details}.

\subsection{Main Results}
\label{sec:main_results}

\noindent\textbf{Comparison with State-of-the-Art Driving World Models.}
On MagicDrive-V2, GenLPO reduces FVD from 94.84 to 78.24 (17.5\%) and FID from 20.91 to 14.54 (30.5\%), achieving the lowest FVD and FID among the compared MagicDrive-V2-based methods (Table~\ref{tab:driving_sota}). GenLPO outperforms RiskMV-DPO~\cite{Lin_2026_CVPR} in both FVD (78.24 vs. 87.65) and FID (14.54 vs. 15.70). Compared with MagicDrive-V2 + Geo~\cite{Jia_2026_CVPR}, GenLPO achieves lower FID (14.54 vs. 18.10) and FVD (78.24 vs. 81.52) without additional geographic-image conditions. GenLPO also improves 3D perception performance, increasing BEVFusion mAP from 17.65 to 18.39 (4.2\%) and NDS from 27.19 to 28.37 (4.3\%).

On UniMLVG, GenLPO reduces FVD from 36.10 to 29.33 (18.8\%) and FID from 5.80 to 5.23 (9.8\%), achieving state-of-the-art video generation quality. These results also surpass UniMLVG + Geo~\cite{Jia_2026_CVPR} (FVD 29.97, FID 5.60), which introduces additional geographic-image conditions. Without introducing additional conditioning inputs, GenLPO further improves BEVFusion mAP from 22.50 to 23.17 (3.0\%) and NDS from 30.16 to 31.09 (3.1\%), achieving the highest BEVFusion mAP and NDS in Table~\ref{tab:driving_sota}. Together, these results demonstrate the effectiveness of generation-informed local preference optimization in improving video quality and driving-scene perception across distinct pretrained driving world models.

\noindent\textbf{Comparison with Post-training Methods.}
As shown in Table~\ref{tab:posttraining}, GenLPO achieves the strongest overall performance among SFT, Vanilla DPO~\cite{Liu_2025_CVPR}, and LocalDPO~\cite{Huang_2026_CVPR} on both MagicDrive-V2 and UniMLVG. Compared with LocalDPO, GenLPO reduces FVD by 10.2\% and 14.0\% and FID by 17.1\% and 5.6\% on the two backbones, respectively, achieving the best video generation quality. These improvements extend to driving-scene perception, with GenLPO achieving the highest BEVFormer mAP (18.75/24.95) and mIoU, BEVFusion NDS (28.37/31.09), and CVT road and vehicle mIoU on both backbones.
For Vanilla DPO, we use real videos as preferred samples and corresponding complete model-generated videos as rejected samples for video-level optimization. LocalDPO selects local regions independently of generated content, whereas GenLPO uses object mismatches and cross-frame geometric inconsistencies to guide region selection and constructs verified local preference pairs from the corresponding model-generated content. These gains show that effective local preference optimization depends not only on supervision granularity, but also on whether the selected regions reflect discrepancies observed in driving videos generated by the current model.

\begin{figure*}[t]
    \centering
    \includegraphics[width=\textwidth]{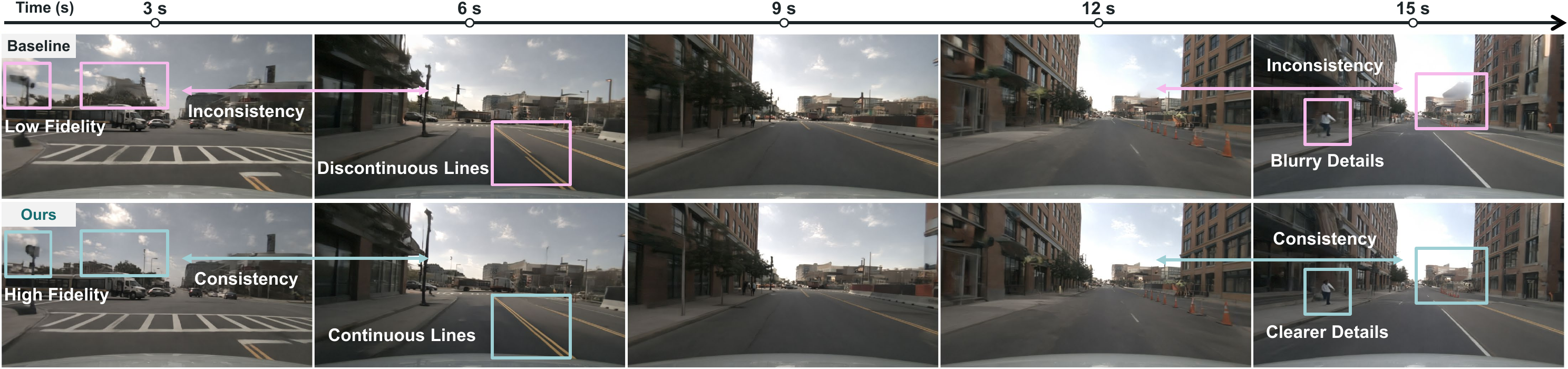}
    \caption{\textbf{Temporal qualitative comparison on MagicDrive-V2.} Highlighted regions show improved temporal consistency, lane-marking continuity, and visual fidelity with GenLPO.}
    \label{fig:temporal_comparison}
\end{figure*}

\GenLPOAblationTable

\subsection{Ablation Studies}
\label{sec:ablation}

We conduct progressive ablations on MagicDrive-V2 to
evaluate the contributions of generation-informed region
localization, preference construction, and iterative
refinement.
The results are summarized in Table~\ref{tab:ablation}.

\begin{figure}[t]
    \centering
    \includegraphics[width=\columnwidth]{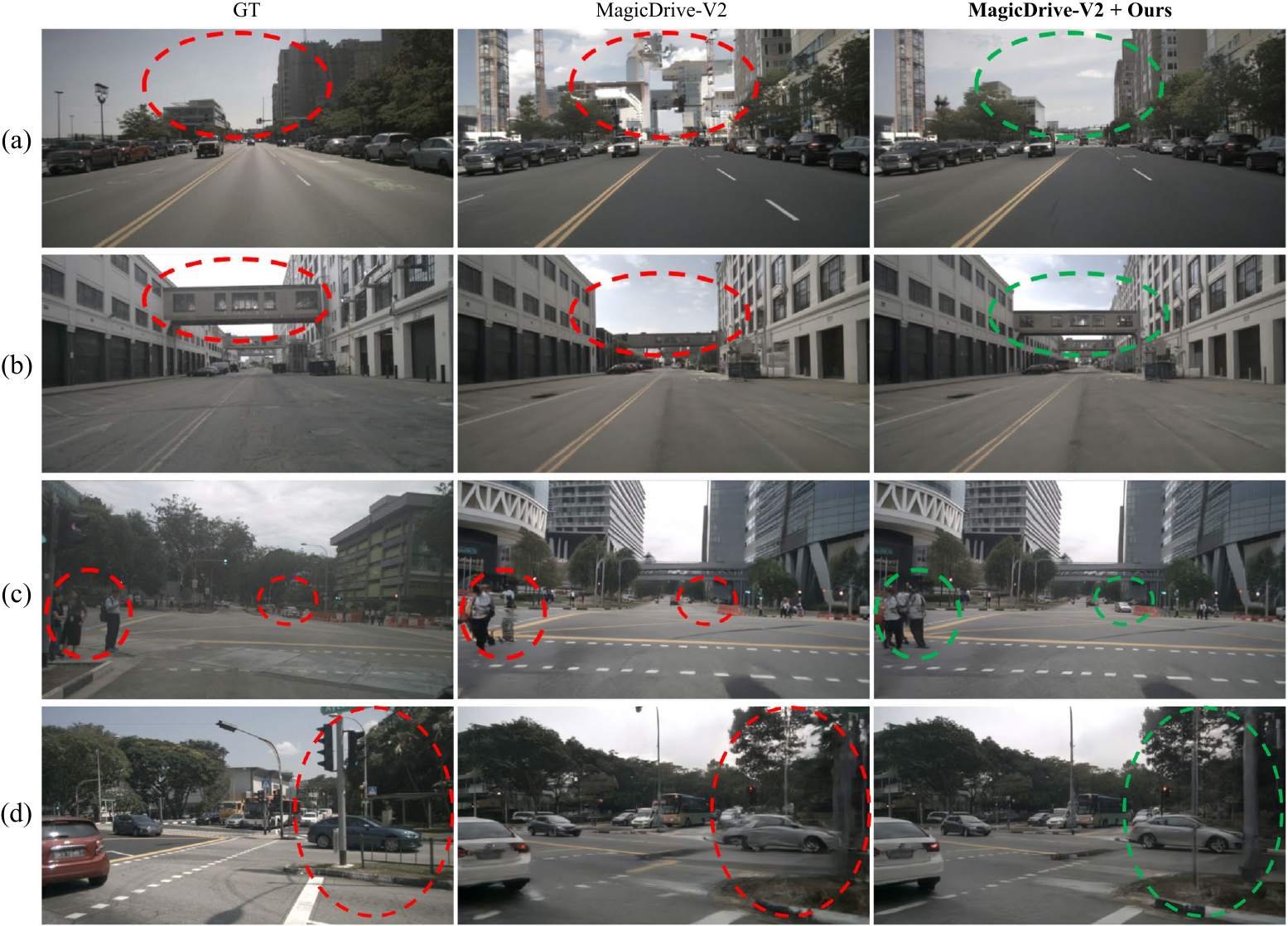}
    \caption{\textbf{Qualitative comparisons of structural and object-level refinements.}
    (a) Spurious Structure Suppression.
    (b) Missing Structure Recovery.
    (c) Missing Object Recovery.
    (d) Vehicle Appearance Refinement.}
    \label{fig:qualitative}
\end{figure}

\noindent\textbf{Generation-Informed Region Localization.}
OML-guided refinement reduces FID from 20.91 to 17.92 and improves NDS from 27.19 to 27.81, demonstrating the benefits of targeting mismatched traffic participants for both visual quality and 3D perception. Incorporating GIL further reduces FVD from 87.71 to 83.93, showing that geometric inconsistencies in static surroundings provide complementary regions for preference optimization.

\noindent\textbf{Generation-Informed Preference Construction.}
With OML and GIL enabled, we compare GIPC against LocalDPO-style local regeneration using the same region localization criteria. GIPC constructs and verifies local preference pairs using model-generated content, directing supervision toward discrepancies that arise during normal generation. This reduces FVD from 83.93 to 81.36 and improves mAP from 18.01 to 18.22, demonstrating the benefit of learning from verified generation discrepancies.

\noindent\textbf{Iterative Preference Refinement.}
Compared with a fixed preference pool under the same training budget, refreshing verified preference pairs using the updated model further reduces FID from 15.77 to 14.54 and FVD from 81.36 to 78.24. These improvements demonstrate the benefit of adapting local preference supervision to newly observed discrepancies throughout post-training.

\subsection{Qualitative Comparisons}
\label{sec:qualitative}

Fig.~\ref{fig:temporal_comparison} shows that GenLPO improves temporal consistency across the generated sequence, with gains extending beyond individual frames.
Fig.~\ref{fig:qualitative} further illustrates how GenLPO refines structural discrepancies in static surroundings (Figs.~\ref{fig:qualitative}(a) and (b)) and object-level discrepancies involving dynamic traffic participants (Figs.~\ref{fig:qualitative}(c) and (d)). These improvements highlight the effectiveness of generation-informed preference supervision in refining both static and dynamic content. Additional qualitative comparisons are provided in Appendix~\ref{supp:qualitative_results}.

%% file: sec/5_conclusion.tex
\section{Conclusion}

We presented GenLPO, a generation-informed local preference optimization framework that uses discrepancies observed in complete model-generated videos to guide region selection and preference construction for pretrained driving world models. OML identifies object mismatches involving dynamic traffic participants, while GIL localizes cross-frame geometric inconsistencies in static surroundings, providing complementary regions for targeted preference supervision. GIPC constructs verified local preference pairs using the corresponding model-generated content, enabling region-aware DPO to learn directly from these discrepancies. The iterative generation-feedback loop refreshes region selection and local preference supervision using videos from the updated model. On nuScenes, GenLPO achieves state-of-the-art video generation quality, reducing FVD by 17.5\% and 18.8\% and FID by 30.5\% and 9.8\% on MagicDrive-V2 and UniMLVG, respectively, while also improving perception-based metrics. Qualitative comparisons further demonstrate improved temporal consistency, clearer scene structures, more complete traffic content, and improved object-level visual fidelity.

%% file: sec/appendix.tex
\clearpage
\maketitlesupplementary
\appendix
\numberwithin{equation}{section}
\raggedbottom

This supplementary material provides additional methodological and implementation details, quantitative analyses, qualitative comparisons, and discussion of GenLPO. The contents are organized as follows:
\begin{itemize}
    \item Sec.~\ref{supp:notation} summarizes the notation used in GenLPO.
    \item Sec.~\ref{supp:region_localization} details object matching, region projection, and geometric consistency computation.
    \item Sec.~\ref{app:gipc_details} describes preference verification and supervision mask construction.
    \item Sec.~\ref{app:iterative_optimization} details the iterative optimization schedule and dynamic-static gradient aggregation.
    \item Sec.~\ref{supp:experimental_details} documents LoRA configurations, optimization settings, and distributed training.
    \item Sec.~\ref{supp:additional_experiments} presents round-wise refinement results and additional object-level and geometric evaluations.
    \item Sec.~\ref{supp:qualitative_results} presents qualitative comparisons of object-level and structural refinements across camera views and time.
    \item Sec.~\ref{supp:discussion} discusses the scope of GenLPO and directions for future work.
\end{itemize}

\section{Notation Table}
\label{supp:notation}

Table~\ref{supp:tab:notation} summarizes the notation used in GenLPO.

\GenLPOSuppNotationTable

\section{Additional Details of Generation-Informed Region Localization}
\label{supp:region_localization}

Building on the localization criteria introduced in Sec.~\ref{sec:region_localization} of the main paper, we further detail the matching and geometric computations underlying OML and GIL. Specifically, for OML, we describe the multi-threshold matching of conditioned 3D boxes to detections in the real video, preferred VAE reconstruction, and generated video, followed by the projection of selected boxes into camera views. For GIL, we elaborate on the cross-frame geometric calculations using depth, optical flow, and camera poses, including the validity conditions that ensure the generated and preferred videos are compared using the same geometric cues and neighboring frames.

\subsection{Object Matching and Region Projection}
\label{supp:object_matching}

To further clarify the OML selection criterion in Sec.~\ref{sec:region_localization} of the main paper, we detail the matching procedure and its behavior across distance thresholds. We apply BEVFormer~\cite{10.1007/978-3-031-20077-9_1} to the real video $\mathbf{X}$, the preferred reconstruction $\mathbf{X}^{+}$, and the generated video $\hat{\mathbf{X}}$, and independently match the detected 3D boxes to the conditioned boxes in each video. Matching uses class-consistent one-to-one associations based on Euclidean distances between box centers in the BEV plane. We use the nuScenes center-distance thresholds $\mathcal{T}=\{0.5,1,2,4\}$\,m~\cite{Caesar_2020_CVPR}. For participant $i$ in frame $t$, $r_{i,t}^{\tau}$, $p_{i,t}^{\tau}$, and $g_{i,t}^{\tau}$ denote binary match indicators in the three videos, respectively, at threshold $\tau\in\mathcal{T}$. To support the following case analysis, we restate the selection criterion in Eq.~\eqref{eq:dynamic_candidates} of the main paper:
\begin{equation}
\label{eq:app_oml_selection}
\begin{aligned}
r_{i,t}^{\tau} &\le p_{i,t}^{\tau}, && \forall \tau\in\mathcal{T},\\
(r_{i,t}^{\tau},p_{i,t}^{\tau},g_{i,t}^{\tau}) &= (1,1,0), && \exists \tau\in\mathcal{T}.
\end{aligned}
\end{equation}
The first condition requires the preferred reconstruction to preserve every real-video match across all thresholds, while allowing additional matches in the reconstruction. The second condition requires at least one threshold at which the real video and preferred reconstruction match but the generated video does not.

Table~\ref{supp:tab:oml_matching} presents five constructed cases illustrating how the two conditions in Eq.~\eqref{eq:app_oml_selection} determine the selection of participant-frame pairs and how selected pairs are classified as \emph{Missing Object} or \emph{Position Shift}. Each case considers one conditioned participant in a single frame and at most one same-class detection candidate per video.

\GenLPOSuppOMLMatchingTable

Cases~1 and~2 satisfy both conditions in Eq.~\eqref{eq:app_oml_selection}, but differ in whether the generated video has a match at $\tau_{\max}=4$\,m. In Case~1, no generated match exists even at $\tau_{\max}$, so the selected pair is classified as \emph{Missing Object}. In Case~2, the distances between the conditioned box center and the detection centers in $\mathbf{X}$, $\mathbf{X}^{+}$, and $\hat{\mathbf{X}}$ are $0.3$, $0.4$, and $1.6$\,m, respectively. Accordingly, the indicator triplet $(r_{i,t}^{\tau},p_{i,t}^{\tau},g_{i,t}^{\tau})$ is $(1,1,0)$ at $0.5$ and $1$\,m, and $(1,1,1)$ at $2$ and $4$\,m. The generated detection matches at $\tau_{\max}$ but not at $0.5$ or $1$\,m, so the pair is classified as \emph{Position Shift}.

Cases~3 and~4 further clarify the asymmetric match-preservation requirement in the first condition of Eq.~\eqref{eq:app_oml_selection}. In Case~3, $(r_{i,t}^{\tau},p_{i,t}^{\tau})=(0,1)$ at $0.5$\,m is permitted by the first condition, and the condition also holds at all other thresholds. The triplet $(1,1,0)$ at $1$\,m satisfies the second condition, so the pair is selected. Since the generated detection matches at $\tau_{\max}$, the pair is classified as \emph{Position Shift}. In Case~4, $(r_{i,t}^{\tau},p_{i,t}^{\tau})=(1,0)$ at $0.5$\,m violates the first condition, so the pair is not selected even though $(1,1,0)$ at $1$\,m satisfies the second condition. Case~5 satisfies only the first condition, as the triplet is $(1,1,1)$ at every threshold.

For each selected participant-frame pair, we project the relevant 3D boxes into each camera view using calibrated camera intrinsics and extrinsics. Perspective division uses $\pi([x,y,z]^{\mathsf T})=[x/z,y/z]^{\mathsf T}$, and the operator $\Pi_{t,v}$ in Eq.~\eqref{eq:dynamic_regions} of the main paper returns the enclosing 2D rectangle of a projected box. For \emph{Missing Object}, we use $\Pi_{t,v}(\mathbf{B}_{i,t})$ for the conditioned box. For \emph{Position Shift}, we take its union with $\Pi_{t,v}(\hat{\mathbf{B}}_{i,t})$, covering both object locations. The union over selected participants yields the per-frame, per-view dynamic candidate regions $\{\Omega_{t,v}^{\mathrm{dyn}}\}_{t,v}$.


\subsection{Geometric Consistency Computation}
\label{sec:app_geometric_consistency}
\label{supp:geometric_consistency}

\noindent\textbf{Cross-frame Geometric Projection.}
For a fixed camera view, we evaluate each video $V\in\{\hat{\mathbf X},\mathbf X^+\}$ separately. To simplify notation, we omit the superscript $V$ in the single-video calculations below and restore it when comparing the two videos. Let $D_t(\mathbf u)$ denote the estimated camera-axis depth at pixel $\mathbf u$ in frame $t$, and let $\mathbf F_{t\rightarrow s}(\mathbf u)$ denote the optical flow to a neighboring frame $s\in\mathcal N(t)$. Following the geometric cues of GeCo~\cite{10.1007/978-3-032-37132-4_28}, we estimate depth using Depth Anything 3~\cite{ICLR2026_e4cd5012} and optical flow using SEA-RAFT~\cite{10.1007/978-3-031-72667-5_3}. 
We use metric-aligned depth estimates from DA3 for geometric reprojection with calibrated camera poses.
Both videos use the same calibrated camera intrinsics and relative poses.

Given intrinsics $\mathbf K_t$ and $\mathbf K_s$ and the relative pose $(\mathbf R_{s\leftarrow t},\mathbf t_{s\leftarrow t})$, we transform the 3D point corresponding to pixel $\mathbf u$ into the target camera frame and compute its rigid projection:
\begin{equation}
\begin{aligned}
\mathbf p_s(\mathbf u)
&=\mathbf R_{s\leftarrow t}\bigl(D_t(\mathbf u)\mathbf K_t^{-1}\tilde{\mathbf u}\bigr)+\mathbf t_{s\leftarrow t},\\
\mathbf u'
&=\pi\bigl(\mathbf K_s\mathbf p_s(\mathbf u)\bigr).
\end{aligned}
\label{eq:app_gil_projection}
\end{equation}
Here $\tilde{\mathbf u}=(u_x,u_y,1)^\top$ is the homogeneous pixel coordinate, $\mathbf p_s(\mathbf u)$ is the corresponding 3D point expressed in the target camera frame, and $\pi([x,y,z]^\top)=(x/z,y/z)^\top$ denotes perspective projection. Under the static-scene assumption, $\mathbf u'$ is the target pixel location predicted from depth and camera motion.

\noindent\textbf{Motion and Structural Consistency Errors.}
Let $\delta\mathbf u=\mathbf u+\mathbf F_{t\rightarrow s}(\mathbf u)-\mathbf u'$ denote the difference between the optical-flow endpoint and the rigid projection, with components $(\delta u_x,\delta u_y)$. The motion consistency error is
\begin{equation}
E_{\mathrm{mot}}(t,s,\mathbf u)
=
\sqrt{
\left(\frac{\delta u_x}{f_{x,t}}\right)^2+
\left(\frac{\delta u_y}{f_{y,t}}\right)^2
},
\label{eq:app_gil_motion}
\end{equation}
where $f_{x,t}$ and $f_{y,t}$ are the focal lengths of the source intrinsics $\mathbf K_t$ along the two image axes.

The structural consistency error compares the transformed source depth with the estimated target depth at the rigid projection:
\begin{equation}
E_{\mathrm{str}}(t,s,\mathbf u)
=
\frac{
\bigl|D_s(\mathbf u')-[\mathbf p_s(\mathbf u)]_z\bigr|
}{
D_t(\mathbf u)
}.
\label{eq:app_gil_structure}
\end{equation}
We sample $D_s$ bilinearly at $\mathbf u'$, and $[\cdot]_z$ denotes the depth component in the target camera frame. The focal-length and source-depth normalizations make the two errors dimensionless.

\noindent\textbf{Geometric Validity Checks.} For each source pixel $\mathbf u$ and neighboring frame $s$, we first exclude known dynamic-object regions and require valid, positive source depth, positive transformed depth, and a rigid projection within the target image. Motion and structure then apply different additional checks. The motion cue requires the optical-flow endpoint $\mathbf u+\mathbf F_{t\rightarrow s}(\mathbf u)$ to lie within the target image and satisfy forward-backward flow consistency. The structural cue instead requires valid target-frame depth at the rigid projection $\mathbf u'$. Since the two cues rely on different target-frame measurements, they are checked independently, allowing either cue to remain valid when the other fails.

These checks are performed separately for $\hat{\mathbf X}$ and $\mathbf X^+$. For the paired comparison, we retain only the cue types valid in both videos at the same pixel and frame pair, forming the set $\mathcal C_{t,s}(\mathbf u)$ already used in Eq.~\eqref{eq:static_fused_error} of the main paper. The resulting errors are averaged over the same valid neighboring frames for both videos, as described in Eq.~\eqref{eq:static_discrepancy}. Pixels without any jointly valid neighboring frame contribute zero to the heatmap in Eq.~\eqref{eq:static_heatmap}.
For region selection, we retain the top 30\% of eligible latent-grid cells ranked by $H_{v,b}(\mathbf q)$ within each camera view and temporal bin, with the number of selected cells rounded down. We then filter connected components with insufficient supporting observations, following Sec.~\ref{sec:region_localization} of the main paper.

\section{GIPC Verification and Mask Construction}
\label{app:gipc_details}

Within each branch, multiple localized discrepancies
can be combined to construct a candidate rejected
sample. GIPC verifies these discrepancies individually
after local replacement and VAE decoding. When only
a subset remains valid, the candidate is refined using
the verified regions and the negative sample is
reconstructed for further verification. For static
surroundings, verification evaluates signed geometric
discrepancies over valid correspondences affected by
the replacement, rather than directly reusing the
nonnegative heatmap scores employed for localization.
A static latent cell passes verification when its mean signed geometric discrepancy over valid correspondences is positive. 
Each retained preference pair is associated with its
final rejected latent and the corresponding verified
regions.

The latent regions used for replacement and preference
supervision are constructed separately. For dynamic
traffic participants, projected object regions define
the candidate edit support, while verified object
events determine the supervision mask $\mathbf{M}$.
For static surroundings, the corresponding supports
are constructed from the selected geometric regions
after verification. The edit support includes
surrounding context, whereas the supervision mask
remains centered on the verified regions. Both are
aligned with the native VAE latent grid of each
backbone~\cite{ICLR2025_ce31378e,pmlr-v235-esser24a}, with the supervision
mask contained within the edit support. Outside the edit support, the
rejected latent remains identical to the preferred
latent.

\section{Iterative Preference Optimization Details}
\label{app:iterative_optimization}

\noindent\textbf{Optimization Schedule.}
We perform four rounds of iterative preference optimization,
with 200 steps in the first round and 100 steps in each
subsequent round, totaling 500 steps.
Starting from the pretrained checkpoint, the first round
establishes the initial preference adaptation.
Subsequent rounds continue from the preceding checkpoint
and use shorter updates to incorporate newly verified
local preference pairs, progressively refining discrepancies
identified by the updated model.

\noindent\textbf{Dynamic-Static Gradient Aggregation.}
At each optimization step, the dynamic and static preference
branches produce gradients $g_{\mathrm{dyn}}$ and
$g_{\mathrm{stat}}$, respectively.
We adopt RACO~\cite{pmlr-v306-chen26bl} to coordinate these
gradients with equal branch weights.
Let $g_0=(g_{\mathrm{dyn}}+g_{\mathrm{stat}})/2$ and
$g(p)=p\,g_{\mathrm{dyn}}+(1-p)\,g_{\mathrm{stat}}$.
The mixing coefficient is obtained by
\begin{equation}
p^*=\arg\min_{p\in[0,1]}
\left[
\langle g(p),g_0\rangle
+\gamma\|g_0\|\|g(p)\|
\right].
\end{equation}
We clip the coefficients as
$\tilde{p}_{\mathrm{dyn}}=\min(p^*,1/2)$ and
$\tilde{p}_{\mathrm{stat}}=\min(1-p^*,1/2)$, and define
$\tilde{g}=\tilde{p}_{\mathrm{dyn}}g_{\mathrm{dyn}}
+\tilde{p}_{\mathrm{stat}}g_{\mathrm{stat}}$.
The combined gradient is
\begin{equation}
g_{\mathrm{joint}}=
\begin{cases}
g_0+\gamma\|g_0\|
\dfrac{\tilde{g}}{\|\tilde{g}\|},
& \|\tilde{g}\|>0,\\[4pt]
g_0, & \text{otherwise}.
\end{cases}
\end{equation}
We set $\gamma=0.4$ and use $g_{\mathrm{joint}}$
to update the shared LoRA adapter~\cite{hu2022lora}.

\section{Additional Implementation Details}
\label{supp:experimental_details}

\noindent\textbf{LoRA Configuration.}
For MagicDrive-V2~\cite{Gao_2025_ICCV}, we apply LoRA to the query,
key, value, and output projections in the spatial self-attention
and cross-attention layers.
This introduces 16.52M trainable parameters.
For UniMLVG~\cite{Chen_2025}, we apply LoRA to the image and text
projections within the joint attention layers, resulting in
18.78M trainable parameters.
All pretrained model weights remain frozen.

\noindent\textbf{Training Setup.}
We use the AdamW optimizer~\cite{Loshchilov_2019_ICLR}
with a weight decay of 0.01.
The learning rate is increased linearly over the first 20 steps,
and gradients are clipped to a maximum norm of 1.0.
Both models are trained on 32 NVIDIA A100 GPUs with gradient
checkpointing enabled.
MagicDrive-V2 uses BF16 precision with sequence parallelism
(SP size 8) and data parallelism (DP size 4), giving a global
batch size of 4 preference pairs per branch.
UniMLVG uses FP16 mixed precision and FSDP with hybrid
sharding~\cite{Zhao_2023_FSDP}, with a global batch size
of 32 preference pairs per branch.
For both models, we set $\beta=1.0$ in the DPO objective
and $\lambda=0.1$ for the preferred-sample flow-matching terms.

\section{Additional Experiments}
\label{supp:additional_experiments}

\subsection{Iterative Refinement Analysis}
\label{supp:iterative_refinement_analysis}

We evaluate FID and FVD at the end of each refinement round
to examine how generation quality evolves as verified local
preferences are refreshed (Sec.~\ref{app:iterative_optimization}).

\GenLPOSuppRoundwiseTable

As shown in Table~\ref{supp:tab:roundwise_optimization},
FID and FVD decrease at every round on both backbones.
From Round 1 to Round 4, MagicDrive-V2 reduces FID from
16.05 to 14.54 (9.41\%) and FVD from 82.60 to 78.24
(5.28\%). On UniMLVG, FID decreases from 5.49 to 5.23
(4.74\%) and FVD from 30.92 to 29.33 (5.14\%).
Together with the ablation in the main paper that keeps the preference pool fixed, these results demonstrate the benefit of refreshing verified local preferences using the updated generator, enabling progressive improvements in image and video generation quality.

\subsection{Additional Evaluation Metrics}
\label{supp:additional_evaluation_metrics}

We further evaluate object-level accuracy and static
geometric consistency to quantify improvements in
dynamic traffic participants and static surroundings.

\noindent\textbf{Object-Level Accuracy.} Using frozen BEVFusion~\cite{Liu_2023}, we report AP@1m, the mean class-wise average precision across ten nuScenes categories at a 1\,m BEV center-distance threshold, and mATE, the class-averaged translation error of matched true positives, averaged over recall levels using the official 2\,m BEV center-distance matching threshold~\cite{Caesar_2020_CVPR}.

\noindent\textbf{Static Geometric Consistency.}
We evaluate the motion and structure errors defined in
Sec.~\ref{supp:geometric_consistency}, using frozen
Depth Anything 3~\cite{ICLR2026_e4cd5012} and
SEA-RAFT~\cite{10.1007/978-3-031-72667-5_3}.
Motion compares optical flow with rigid flow computed from depth and camera pose, normalizing each residual component by its corresponding focal length.
Structure measures the absolute
difference between reprojected depth and estimated
target-frame depth, normalized by source depth.
We exclude annotated traffic-participant regions, sky,
and invalid correspondences, and compute each metric
over pixels jointly valid in the pretrained and GenLPO
videos. The evaluation covers both directions of
adjacent-frame pairs across all 16 generated frames,
six camera views, and 150 validation scenes. Errors
are averaged within each camera sequence and then
aggregated with equal weighting across scenes.

\GenLPOSuppAdditionalMetricsTable

\noindent\textbf{Results Analysis.}
As shown in Table~\ref{supp:tab:additional_metrics},
GenLPO consistently improves object-level accuracy.
On MagicDrive-V2, AP@1m increases from 11.85\% to
12.49\% (5.40\%), while mATE decreases from 0.853
to 0.837\,m (1.88\%). On UniMLVG, AP@1m improves
from 11.34\% to 11.95\% (5.38\%), and mATE decreases
from 0.866 to 0.845\,m (2.42\%).

For static geometric consistency, GenLPO reduces
motion and structure errors from 0.0160 to 0.0143
(10.63\%) and from 0.0416 to 0.0388 (6.73\%) on
MagicDrive-V2, respectively. On UniMLVG, the
corresponding errors decrease from 0.0119 to 0.0107
(10.08\%) and from 0.0407 to 0.0383 (5.90\%).
These consistent improvements demonstrate that GenLPO
enhances both object-level spatial accuracy and
cross-frame geometric consistency, further validating
generation-informed local preference optimization.

\section{Additional Qualitative Results}
\label{supp:qualitative_results}

\begin{figure*}[!t]
\centering
\includegraphics[width=\textwidth]{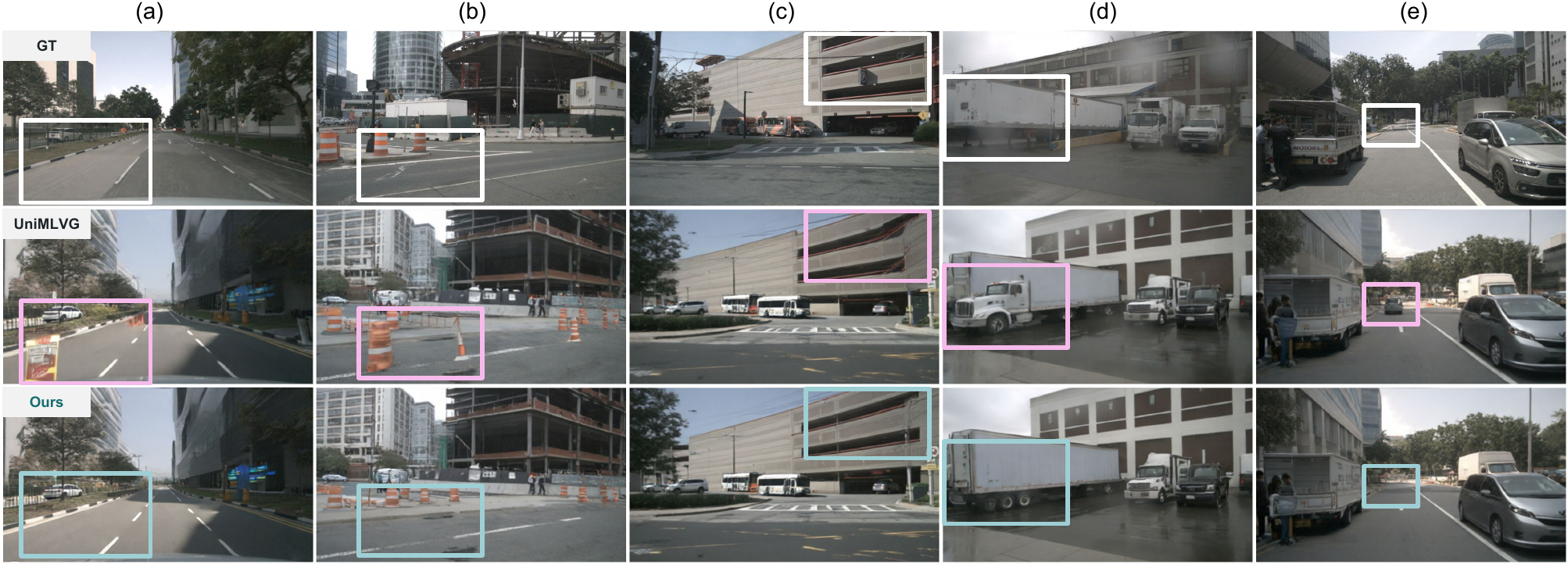}
\caption{\textbf{Qualitative comparison on UniMLVG.}
(a) Spurious Roadside Object. (b) Misplaced Traffic Barrels. (c) Distorted Building Facade. (d) Spurious Truck Cab. 
(e) Spurious Vehicle.}
\label{fig:supp_unimlvg_qualitative}
\end{figure*}

\begin{figure*}[!t]
\centering
\includegraphics[width=\textwidth]{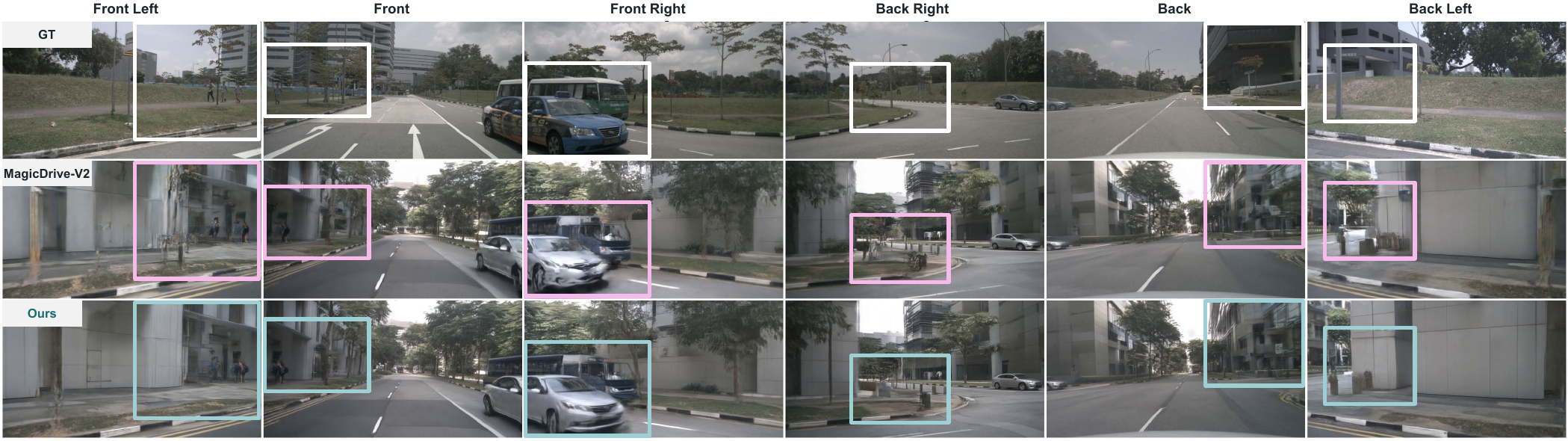}
\caption{\textbf{Multi-view qualitative comparison on MagicDrive-V2.} GenLPO improves overall visual fidelity across six camera views, with clearer vehicle appearance and roadside structures. Boxes highlight representative local differences.}
\label{fig:supp_magicdrive_multiview}
\end{figure*}

Figure~\ref{fig:supp_unimlvg_qualitative} presents representative object- and structure-level refinements on UniMLVG. GenLPO suppresses spurious roadside objects and vehicles, corrects misplaced traffic barrels, and improves the structural continuity of building facades. These examples illustrate improvements in object presence, spatial placement, and static scene structure. The six-view comparison on MagicDrive-V2 in Fig.~\ref{fig:supp_magicdrive_multiview} further demonstrates clearer vehicle appearance and more coherent roadside structures across different camera directions, highlighting the benefits of local refinement within complete multi-view scenes.

\begin{figure*}[p]
\centering
\includegraphics[width=\textwidth,height=0.94\textheight,keepaspectratio]{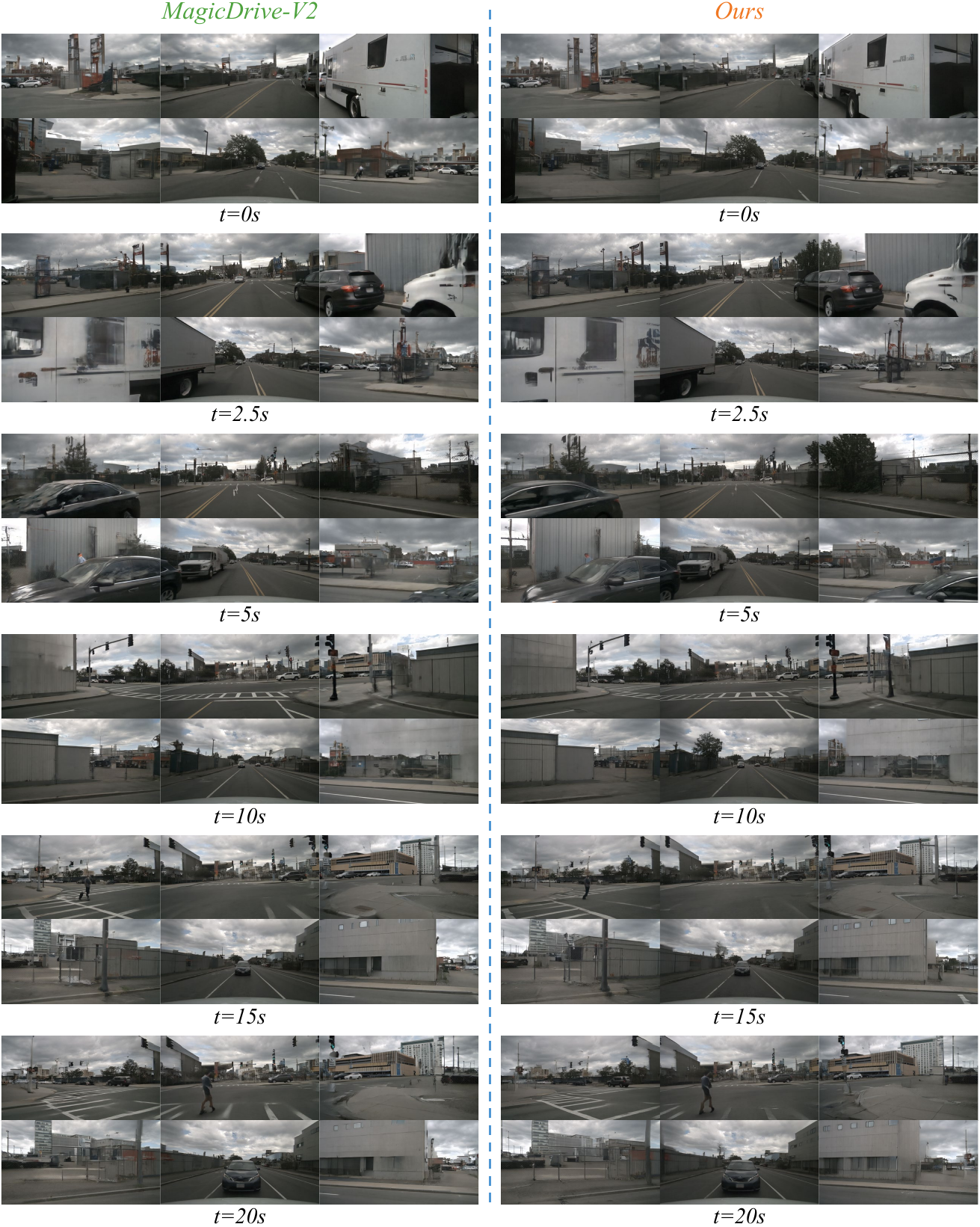}
\caption{\textbf{Temporal qualitative comparison on MagicDrive-V2 (Scene 1).}
Across a 20-second sequence, GenLPO improves the temporal continuity of
roadside structures and surrounding traffic across six camera views.}
\label{fig:supp_magicdrive_temporal}
\end{figure*}

\begin{figure*}[p]
\centering
\includegraphics[width=\textwidth,height=0.94\textheight,keepaspectratio]{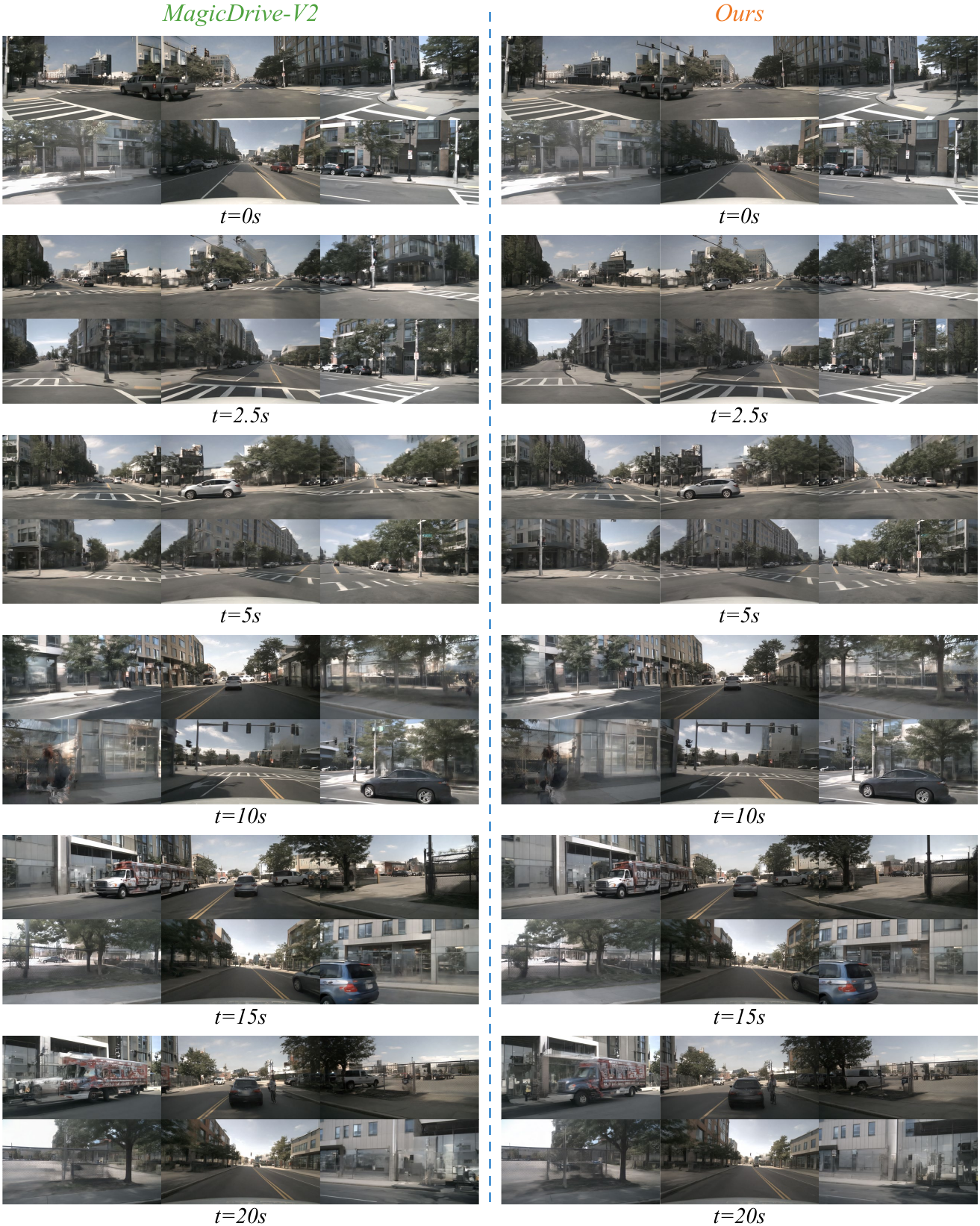}
\caption{\textbf{Temporal qualitative comparison on MagicDrive-V2 (Scene 2).}
Across a 20-second urban driving sequence, GenLPO improves road layout
continuity and the visual fidelity of traffic participants across six
camera views.}
\label{fig:supp_magicdrive_temporal_additional}
\end{figure*}

The two 20-second driving sequences in Figs.~\ref{fig:supp_magicdrive_temporal} and~\ref{fig:supp_magicdrive_temporal_additional} further illustrate how these improvements extend over time. In the first scene, roadside structures and surrounding traffic exhibit greater visual continuity as the viewpoint changes. In the second scene, GenLPO improves road layout continuity and the visual fidelity of traffic participants throughout the urban driving sequence. Together, these comparisons demonstrate improvements in object-level visual fidelity, scene structures, and temporal consistency across camera views and time.

\section{Discussion}
\label{supp:discussion}

GenLPO highlights the importance of considering region selection and preference construction together in local preference optimization. Specifically, region selection identifies the content to refine, whereas preference construction specifies the comparison that guides refinement. By grounding both in discrepancies observed in complete model-generated videos, GenLPO establishes a direct connection between the regions requiring refinement and the supervision used to refine them. Importantly, verification preserves this connection by confirming that the targeted discrepancies persist in the constructed negative samples.
Consistent with this perspective, the preference-construction ablation further demonstrates the value of verified generation-informed pairs under the same localization criteria.

More broadly, this perspective connects scene understanding with the refinement of pretrained driving world models. In particular, object correspondence and geometric coherence provide complementary forms of scene feedback, yet both can be translated into a shared local preference formulation. Building on this principle, the same formulation could be extended to additional scene properties, such as object identity consistency across camera views and longer sequences, through corresponding localization and verification criteria. Future work can further explore these extensions across diverse driving environments and pretrained models.

\makeatletter
\setlength{\@dblfptop}{0pt}
\setlength{\@dblfpsep}{\dblfloatsep}
\makeatother
\clearpage

%% file: main.bbl
\begin{thebibliography}{55}
\providecommand{\natexlab}[1]{#1}
\providecommand{\url}[1]{\texttt{#1}}
\expandafter\ifx\csname urlstyle\endcsname\relax
  \providecommand{\doi}[1]{doi: #1}\else
  \providecommand{\doi}{doi: \begingroup \urlstyle{rm}\Url}\fi

\bibitem[Black et~al.(2024)Black, Janner, Du, Kostrikov, and Levine]{ICLR2024_14f75513}
Kevin Black, Michael Janner, Yilun Du, Ilya Kostrikov, and Sergey Levine.
\newblock Training diffusion models with reinforcement learning.
\newblock In \emph{International Conference on Learning Representations}, pages 4965--4987, 2024.

\bibitem[Caesar et~al.(2020)Caesar, Bankiti, Lang, Vora, Liong, Xu, Krishnan, Pan, Baldan, and Beijbom]{Caesar_2020_CVPR}
Holger Caesar, Varun Bankiti, Alex~H. Lang, Sourabh Vora, Venice~Erin Liong, Qiang Xu, Anush Krishnan, Yu Pan, Giancarlo Baldan, and Oscar Beijbom.
\newblock nuscenes: A multimodal dataset for autonomous driving.
\newblock In \emph{Proceedings of the IEEE/CVF Conference on Computer Vision and Pattern Recognition (CVPR)}, 2020.

\bibitem[Chen et~al.(2025{\natexlab{a}})Chen, Jiang, Liang, Zhuang, Su, Garg, Wu, and Chandraker]{Chen_2025_ICCV}
Jiacheng Chen, Ziyu Jiang, Mingfu Liang, Bingbing Zhuang, Jong-Chyi Su, Sparsh Garg, Ying Wu, and Manmohan Chandraker.
\newblock Autoscape: Geometry-consistent long-horizon scene generation.
\newblock In \emph{Proceedings of the IEEE/CVF International Conference on Computer Vision (ICCV)}, pages 25700--25711, 2025{\natexlab{a}}.

\bibitem[Chen et~al.(2026)Chen, Li, Chen, and Lin]{pmlr-v306-chen26bl}
Peter Chen, Xiaopeng Li, Xi Chen, and Tianyi Lin.
\newblock Reward-free alignment for conflicting objectives.
\newblock In \emph{Proceedings of the 43rd International Conference on Machine Learning}, pages 15091--15117. PMLR, 2026.

\bibitem[Chen et~al.(2025{\natexlab{b}})Chen, Wu, Liu, Guo, Ni, Xia, and Xia]{Chen_2025}
Rui Chen, Zehuan Wu, Yichen Liu, Yuxin Guo, Jingcheng Ni, Haifeng Xia, and Siyu Xia.
\newblock Unimlvg: Unified framework for multi-view long video generation with comprehensive control capabilities for autonomous driving.
\newblock In \emph{Proceedings of the IEEE/CVF International Conference on Computer Vision (ICCV)}, pages 25453--25463, 2025{\natexlab{b}}.

\bibitem[Cheng et~al.(2025)Cheng, Yang, Huang, Si, Si, and Liu]{cheng2025realdporealrealpreference}
Guo Cheng, Danni Yang, Ziqi Huang, Jianlou Si, Chenyang Si, and Ziwei Liu.
\newblock Realdpo: Real or not real, that is the preference.
\newblock \emph{arXiv preprint arXiv:2510.14955}, 2025.

\bibitem[Dong et~al.(2025)Dong, Wang, Lin, Wu, Chen, Liu, Yang, Li, and Guo]{Dong_2025_ICCV}
Haotian Dong, Xin Wang, Di Lin, Yipeng Wu, Qin Chen, Ruonan Liu, Kairui Yang, Ping Li, and Qing Guo.
\newblock Noisecontroller: Towards consistent multi-view video generation via noise decomposition and collaboration.
\newblock In \emph{Proceedings of the IEEE/CVF International Conference on Computer Vision (ICCV)}, pages 14443--14452, 2025.

\bibitem[Esser et~al.(2024)Esser, Kulal, Blattmann, Entezari, M\"{u}ller, Saini, Levi, Lorenz, Sauer, Boesel, Podell, Dockhorn, English, and Rombach]{pmlr-v235-esser24a}
Patrick Esser, Sumith Kulal, Andreas Blattmann, Rahim Entezari, Jonas M\"{u}ller, Harry Saini, Yam Levi, Dominik Lorenz, Axel Sauer, Frederic Boesel, Dustin Podell, Tim Dockhorn, Zion English, and Robin Rombach.
\newblock Scaling rectified flow transformers for high-resolution image synthesis.
\newblock In \emph{Proceedings of the 41st International Conference on Machine Learning}, pages 12606--12633. PMLR, 2024.

\bibitem[Gao et~al.(2024)Gao, Chen, Xie, HONG, Li, Yeung, and Xu]{ICLR2024_635351dc}
Ruiyuan Gao, Kai Chen, Enze Xie, Lanqing HONG, Zhenguo Li, Dit-Yan Yeung, and Qiang Xu.
\newblock Magicdrive: Street view generation with diverse 3d geometry control.
\newblock In \emph{International Conference on Learning Representations}, pages 22841--22860, 2024.

\bibitem[Gao et~al.(2025)Gao, Chen, Xiao, Hong, Li, and Xu]{Gao_2025_ICCV}
Ruiyuan Gao, Kai Chen, Bo Xiao, Lanqing Hong, Zhenguo Li, and Qiang Xu.
\newblock Magicdrive-v2: High-resolution long video generation for autonomous driving with adaptive control.
\newblock In \emph{Proceedings of the IEEE/CVF International Conference on Computer Vision (ICCV)}, pages 28135--28144, 2025.

\bibitem[Gu et~al.(2026)Gu, Hur, Herrmann, Zhan, Zickler, Sun, and Pfister]{10.1007/978-3-032-37132-4_28}
Leslie Gu, Junhwa Hur, Charles Herrmann, Fangneng Zhan, Todd Zickler, Deqing Sun, and Hanspeter Pfister.
\newblock Geco: Evaluating geometric consistency for video generation via motion and structure.
\newblock In \emph{Computer Vision -- ECCV 2026}, pages 499--517, Cham, 2026. Springer Nature Switzerland.

\bibitem[Guo et~al.(2025)Guo, Ding, Chen, Chen, Li, Zou, Lyu, Tan, Qi, Li, and Zhao]{Guo_2025_ICCV}
Jiazhe Guo, Yikang Ding, Xiwu Chen, Shuo Chen, Bohan Li, Yingshuang Zou, Xiaoyang Lyu, Feiyang Tan, Xiaojuan Qi, Zhiheng Li, and Hao Zhao.
\newblock Dist-4d: Disentangled spatiotemporal diffusion with metric depth for 4d driving scene generation.
\newblock In \emph{Proceedings of the IEEE/CVF International Conference on Computer Vision (ICCV)}, pages 27231--27241, 2025.

\bibitem[Heusel et~al.(2017)Heusel, Ramsauer, Unterthiner, Nessler, and Hochreiter]{NIPS2017_8a1d6947}
Martin Heusel, Hubert Ramsauer, Thomas Unterthiner, Bernhard Nessler, and Sepp Hochreiter.
\newblock Gans trained by a two time-scale update rule converge to a local nash equilibrium.
\newblock In \emph{Advances in Neural Information Processing Systems}. Curran Associates, Inc., 2017.

\bibitem[Hu et~al.(2023)Hu, Russell, Yeo, Murez, Fedoseev, Kendall, Shotton, and Corrado]{hu2023gaia1generativeworldmodel}
Anthony Hu, Lloyd Russell, Hudson Yeo, Zak Murez, George Fedoseev, Alex Kendall, Jamie Shotton, and Gianluca Corrado.
\newblock Gaia-1: A generative world model for autonomous driving, 2023.

\bibitem[Hu et~al.(2022)Hu, Shen, Wallis, Allen-Zhu, Li, Wang, Wang, and Chen]{hu2022lora}
Edward~J Hu, Yelong Shen, Phillip Wallis, Zeyuan Allen-Zhu, Yuanzhi Li, Shean Wang, Lu Wang, and Weizhu Chen.
\newblock Lo{RA}: Low-rank adaptation of large language models.
\newblock In \emph{International Conference on Learning Representations}, 2022.

\bibitem[Huang et~al.(2026)Huang, Zhang, Ding, Gao, Ding, Chen, and Zuo]{Huang_2026_CVPR}
Zitong Huang, Kaidong Zhang, Yukang Ding, Chao Gao, Rui Ding, Ying Chen, and Wangmeng Zuo.
\newblock Mind the generative details: Direct localized detail preference optimization for video diffusion models.
\newblock In \emph{Proceedings of the IEEE/CVF Conference on Computer Vision and Pattern Recognition (CVPR)}, pages 35998--36008, 2026.

\bibitem[Jia et~al.(2026)Jia, Zhang, Jiang, Wong, Zhang, Chen, Zhang, Zhou, Yang, Yan, and Jiang]{Jia_2026_CVPR}
Xiaosong Jia, Chenhe Zhang, Yule Jiang, Songbur Wong, Zhiyuan Zhang, Chen Chen, Shaofeng Zhang, Xuanhe Zhou, Xue Yang, Junchi Yan, and Yu-Gang Jiang.
\newblock Spatial retrieval augmented autonomous driving.
\newblock In \emph{Proceedings of the IEEE/CVF Conference on Computer Vision and Pattern Recognition (CVPR)}, pages 17787--17797, 2026.

\bibitem[Jiang et~al.(2026)Jiang, Hong, Zhang, Hu, Zhan, Shao, and Nie]{Jiang_2026}
Junpeng Jiang, Gangyi Hong, Miao Zhang, Hengtong Hu, Kun Zhan, Rui Shao, and Liqiang Nie.
\newblock Dive: Efficient multi-view driving scenes generation based on video diffusion transformer.
\newblock \emph{IEEE Transactions on Multimedia}, 28:\penalty0 11903–11917, 2026.

\bibitem[Li et~al.(2025{\natexlab{a}})Li, Long, Zheng, Gao, Piramuthu, Chen, and Wang]{ICLR2025_e68af7d8}
Jiachen Li, Qian Long, Jian~(Skyler) Zheng, Xiaofeng Gao, Robinson Piramuthu, Wenhu Chen, and William Wang.
\newblock T2v-turbo-v2: Enhancing video model post-training through data, reward, and conditional guidance design.
\newblock In \emph{International Conference on Learning Representations}, pages 92279--92305, 2025{\natexlab{a}}.

\bibitem[Li et~al.(2026{\natexlab{a}})Li, Wang, Gao, Liu, Li, An, and Guo]{Li_2026_CVPR}
Jiaxing Li, Jiepeng Wang, Junyao Gao, Yang Liu, Eric Li, Bo An, and Hao-Xiang Guo.
\newblock Dynamicsboost: Dynamic plausible video generation via annotation-free continuation preference optimization.
\newblock In \emph{Proceedings of the IEEE/CVF Conference on Computer Vision and Pattern Recognition (CVPR)}, pages 20024--20033, 2026{\natexlab{a}}.

\bibitem[Li et~al.(2026{\natexlab{b}})Li, Xu, Tseng, Lu, Liu, and Lan]{li2026lineardpolineardirectpreference}
Kesong Li, Yixuan Xu, {Kuo-kun} Tseng, Weiyi Lu, Kan Liu, and Tao Lan.
\newblock Linear-dpo: Linear direct preference optimization for diffusion and flow-matching generative models, 2026{\natexlab{b}}.

\bibitem[Li et~al.(2025{\natexlab{b}})Li, Zhang, and Ye]{10.1007/978-3-031-73229-4_27}
Xiaofan Li, Yifu Zhang, and Xiaoqing Ye.
\newblock Drivingdiffusion: Layout-guided multi-view driving scenarios video generation with latent diffusion model.
\newblock In \emph{Computer Vision -- ECCV 2024}, pages 469--485, Cham, 2025{\natexlab{b}}. Springer Nature Switzerland.

\bibitem[Li et~al.(2022)Li, Wang, Li, Xie, Sima, Lu, Qiao, and Dai]{10.1007/978-3-031-20077-9_1}
Zhiqi Li, Wenhai Wang, Hongyang Li, Enze Xie, Chonghao Sima, Tong Lu, Yu Qiao, and Jifeng Dai.
\newblock Bevformer: Learning bird's-eye-view representation from multi-camera images via spatiotemporal transformers.
\newblock In \emph{Computer Vision -- ECCV 2022}, pages 1--18, Cham, 2022. Springer Nature Switzerland.

\bibitem[Liang et~al.(2026)Liang, Kong, Yan, Liu, Yang, Huang, Yin, Zuo, Hu, Zhu, Lu, Liu, Jiang, Li, Li, Zhuo, Ng, Cottereau, Gao, Pan, Ooi, and Liu]{Liang_2026_CVPR}
Ao Liang, Lingdong Kong, Tianyi Yan, Hongsi Liu, Yu Yang, Ziqi Huang, Wei Yin, Jialong Zuo, Yixuan Hu, Dekai Zhu, Dongyue Lu, Youquan Liu, Guangfeng Jiang, Linfeng Li, Xiangtai Li, Long Zhuo, Lai~Xing Ng, Benoit~R. Cottereau, Changxin Gao, Liang Pan, Wei~Tsang Ooi, and Ziwei Liu.
\newblock Worldlens: Full-spectrum evaluations of driving world models in real world.
\newblock In \emph{Proceedings of the IEEE/CVF Conference on Computer Vision and Pattern Recognition (CVPR)}, pages 36385--36399, 2026.

\bibitem[Lin et~al.(2026{\natexlab{a}})Lin, Chen, Liew, Chen, Li, Zhao, Peng, Guo, Zhou, Shi, Feng, and Kang]{ICLR2026_e4cd5012}
Haotong Lin, Sili Chen, Jun~Hao Liew, Donny~Y. Chen, Zhenyu Li, Yang Zhao, Sida Peng, Hengkai Guo, Xiaowei Zhou, Guang Shi, Jiashi Feng, and Bingyi Kang.
\newblock Depth anything 3: Recovering the visual space from any views.
\newblock In \emph{International Conference on Learning Representations}, pages 141261--141285, 2026{\natexlab{a}}.

\bibitem[Lin et~al.(2026{\natexlab{b}})Lin, Shi, Huang, Zhuang, Zhang, Liu, Qu, and Zhao]{Lin_2026_CVPR}
Hongyi Lin, Wenxiu Shi, Heye Huang, Dingyi Zhuang, Song Zhang, Yang Liu, Xiaobo Qu, and Jinhua Zhao.
\newblock Risk-controllable multi-view diffusion for driving scenario generation.
\newblock In \emph{Proceedings of the IEEE/CVF Conference on Computer Vision and Pattern Recognition (CVPR) Workshops}, pages 5169--5178, 2026{\natexlab{b}}.

\bibitem[Liu et~al.(2025{\natexlab{a}})Liu, Liu, Liang, Yuan, Liu, Zheng, Wu, Wang, Xia, Wang, Liu, Yang, Wan, ZHANG, Gai, Yang, and Ouyang]{NEURIPS2025_76227feb}
Jie Liu, Gongye Liu, Jiajun Liang, Ziyang Yuan, Xiaokun Liu, Mingwu Zheng, Xiele Wu, Qiulin Wang, Menghan Xia, Xintao Wang, Xiaohong Liu, Fei Yang, Pengfei Wan, Di ZHANG, Kun Gai, Yujiu Yang, and Wanli Ouyang.
\newblock Improving video generation with human feedback.
\newblock In \emph{Advances in Neural Information Processing Systems}, pages 82155--82192. Curran Associates, Inc., 2025{\natexlab{a}}.

\bibitem[Liu et~al.(2026)Liu, Li, Zhang, Li, Yu, Fan, Zheng, and Bai]{10.1007/978-3-032-37531-5_22}
Jiazheng Liu, Hang Li, Jiawei Zhang, Jiahe Li, Xiaohan Yu, Shengyin Fan, Jin Zheng, and Xiao Bai.
\newblock Geoflow: Efficient driving video generation via geometry-aligned priors.
\newblock In \emph{Computer Vision -- ECCV 2026}, pages 394--412, Cham, 2026. Springer Nature Switzerland.

\bibitem[Liu et~al.(2025{\natexlab{b}})Liu, Wu, Zheng, Wei, He, Pi, and Chen]{Liu_2025_CVPR}
Runtao Liu, Haoyu Wu, Ziqiang Zheng, Chen Wei, Yingqing He, Renjie Pi, and Qifeng Chen.
\newblock Videodpo: Omni-preference alignment for video diffusion generation.
\newblock In \emph{Proceedings of the IEEE/CVF Conference on Computer Vision and Pattern Recognition (CVPR)}, pages 8009--8019, 2025{\natexlab{b}}.

\bibitem[Liu et~al.(2023)Liu, Tang, Amini, Yang, Mao, Rus, and Han]{Liu_2023}
Zhijian Liu, Haotian Tang, Alexander Amini, Xinyu Yang, Huizi Mao, Daniela~L. Rus, and Song Han.
\newblock Bevfusion: Multi-task multi-sensor fusion with unified bird's-eye view representation.
\newblock In \emph{2023 IEEE International Conference on Robotics and Automation (ICRA)}, pages 2774--2781, 2023.

\bibitem[Loshchilov and Hutter(2019)]{Loshchilov_2019_ICLR}
Ilya Loshchilov and Frank Hutter.
\newblock Decoupled weight decay regularization.
\newblock In \emph{International Conference on Learning Representations}, 2019.

\bibitem[Lu et~al.(2025)Lu, Huang, Yang, Zhang, and Zhang]{10.1007/978-3-031-72989-8_19}
Jiachen Lu, Ze Huang, Zeyu Yang, Jiahui Zhang, and Li Zhang.
\newblock Wovogen: World volume-aware diffusion for controllable multi-camera driving scene generation.
\newblock In \emph{Computer Vision -- ECCV 2024}, pages 329--345, Cham, 2025. Springer Nature Switzerland.

\bibitem[Mei et~al.(2025)Mei, Hu, Yang, Wen, Yang, Wei, Ma, Dou, Shi, and Liu]{mei2025dreamforgemotionawareautoregressivevideo}
Jianbiao Mei, Tao Hu, Xuemeng Yang, Licheng Wen, Yu Yang, Tiantian Wei, Yukai Ma, Min Dou, Botian Shi, and Yong Liu.
\newblock Dreamforge: Motion-aware autoregressive video generation for multi-view driving scenes, 2025.

\bibitem[Ni et~al.(2025)Ni, Guo, Liu, Chen, Lu, and Wu]{Ni_2025_CVPR}
Jingcheng Ni, Yuxin Guo, Yichen Liu, Rui Chen, Lewei Lu, and Zehuan Wu.
\newblock Maskgwm: A generalizable driving world model with video mask reconstruction.
\newblock In \emph{Proceedings of the IEEE/CVF Conference on Computer Vision and Pattern Recognition (CVPR)}, pages 22381--22391, 2025.

\bibitem[Prabhudesai et~al.(2024)Prabhudesai, Mendonca, Qin, Fragkiadaki, and Pathak]{prabhudesai2024videodiffusionalignmentreward}
Mihir Prabhudesai, Russell Mendonca, Zheyang Qin, Katerina Fragkiadaki, and Deepak Pathak.
\newblock Video diffusion alignment via reward gradients, 2024.

\bibitem[Rafailov et~al.(2023)Rafailov, Sharma, Mitchell, Manning, Ermon, and Finn]{NEURIPS2023_a85b405e}
Rafael Rafailov, Archit Sharma, Eric Mitchell, Christopher~D Manning, Stefano Ermon, and Chelsea Finn.
\newblock Direct preference optimization: Your language model is secretly a reward model.
\newblock In \emph{Advances in Neural Information Processing Systems}, pages 53728--53741. Curran Associates, Inc., 2023.

\bibitem[Unterthiner et~al.(2018)Unterthiner, van Steenkiste, Kurach, Marinier, Michalski, and Gelly]{unterthiner2018towards}
Thomas Unterthiner, Sjoerd van Steenkiste, Karol Kurach, Raphael Marinier, Marcin Michalski, and Sylvain Gelly.
\newblock Towards accurate generative models of video: A new metric \& challenges.
\newblock \emph{arXiv preprint arXiv:1812.01717}, 2018.

\bibitem[Wallace et~al.(2024)Wallace, Dang, Rafailov, Zhou, Lou, Purushwalkam, Ermon, Xiong, Joty, and Naik]{Wallace_2024_CVPR}
Bram Wallace, Meihua Dang, Rafael Rafailov, Linqi Zhou, Aaron Lou, Senthil Purushwalkam, Stefano Ermon, Caiming Xiong, Shafiq Joty, and Nikhil Naik.
\newblock Diffusion model alignment using direct preference optimization.
\newblock In \emph{Proceedings of the IEEE/CVF Conference on Computer Vision and Pattern Recognition (CVPR)}, pages 8228--8238, 2024.

\bibitem[Wang et~al.(2025{\natexlab{a}})Wang, Zhu, Huang, Chen, Zhu, and Lu]{10.1007/978-3-031-73195-2_4}
Xiaofeng Wang, Zheng Zhu, Guan Huang, Xinze Chen, Jiagang Zhu, and Jiwen Lu.
\newblock Drivedreamer: Towards real-world-drive world models for autonomous driving.
\newblock In \emph{Computer Vision -- ECCV 2024}, pages 55--72, Cham, 2025{\natexlab{a}}. Springer Nature Switzerland.

\bibitem[Wang et~al.(2024)Wang, He, Fan, Li, Chen, and Zhang]{Wang_2024_CVPR}
Yuqi Wang, Jiawei He, Lue Fan, Hongxin Li, Yuntao Chen, and Zhaoxiang Zhang.
\newblock Driving into the future: Multiview visual forecasting and planning with world model for autonomous driving.
\newblock In \emph{Proceedings of the IEEE/CVF Conference on Computer Vision and Pattern Recognition (CVPR)}, pages 14749--14759, 2024.

\bibitem[Wang et~al.(2025{\natexlab{b}})Wang, Lipson, and Deng]{10.1007/978-3-031-72667-5_3}
Yihan Wang, Lahav Lipson, and Jia Deng.
\newblock Sea-raft: Simple, efficient, accurate raft for optical flow.
\newblock In \emph{Computer Vision -- ECCV 2024}, pages 36--54, Cham, 2025{\natexlab{b}}. Springer Nature Switzerland.

\bibitem[Wen et~al.(2024)Wen, Zhao, Liu, Jia, Wang, Luo, Zhang, Wang, Sun, and Zhang]{Wen_2024_CVPR}
Yuqing Wen, Yucheng Zhao, Yingfei Liu, Fan Jia, Yanhui Wang, Chong Luo, Chi Zhang, Tiancai Wang, Xiaoyan Sun, and Xiangyu Zhang.
\newblock Panacea: Panoramic and controllable video generation for autonomous driving.
\newblock In \emph{Proceedings of the IEEE/CVF Conference on Computer Vision and Pattern Recognition (CVPR)}, pages 6902--6912, 2024.

\bibitem[Wu et~al.(2025{\natexlab{a}})Wu, Guo, Tang, Huang, Wang, and Ding]{Wu_2025_CVPR}
Wei Wu, Xi Guo, Weixuan Tang, Tingxuan Huang, Chiyu Wang, and Chenjing Ding.
\newblock Drivescape: High-resolution driving video generation by multi-view feature fusion.
\newblock In \emph{Proceedings of the IEEE/CVF Conference on Computer Vision and Pattern Recognition (CVPR)}, pages 17187--17196, 2025{\natexlab{a}}.

\bibitem[Wu et~al.(2025{\natexlab{b}})Wu, Kag, Skorokhodov, Menapace, Mirzaei, Gilitschenski, Tulyakov, and Siarohin]{NEURIPS2025_fa975504}
Ziyi Wu, Anil Kag, Ivan Skorokhodov, Willi Menapace, Ashkan Mirzaei, Igor Gilitschenski, Sergey Tulyakov, and Aliaksandr Siarohin.
\newblock Densedpo: Fine-grained temporal preference optimization for video diffusion models.
\newblock In \emph{Advances in Neural Information Processing Systems}, pages 171632--171668. Curran Associates, Inc., 2025{\natexlab{b}}.

\bibitem[Yan et~al.(2025)Yan, Han, zhou, Zhang, Zhan, Xu, and Shen]{NEURIPS2025_bb0f9af6}
Tianyi Yan, Wencheng Han, xia zhou, Xueyang Zhang, Kun Zhan, Cheng-Zhong Xu, and Jianbing Shen.
\newblock Rlgf: Reinforcement learning with geometric feedback for autonomous driving video generation.
\newblock In \emph{Advances in Neural Information Processing Systems}, pages 128659--128684. Curran Associates, Inc., 2025.

\bibitem[Yang et~al.(2024)Yang, Tao, Lyu, Ge, Chen, Shen, Zhu, and Li]{Yang_2024_CVPR}
Kai Yang, Jian Tao, Jiafei Lyu, Chunjiang Ge, Jiaxin Chen, Weihan Shen, Xiaolong Zhu, and Xiu Li.
\newblock Using human feedback to fine-tune diffusion models without any reward model.
\newblock In \emph{Proceedings of the IEEE/CVF Conference on Computer Vision and Pattern Recognition (CVPR)}, pages 8941--8951, 2024.

\bibitem[Yang et~al.(2025{\natexlab{a}})Yang, Guo, Ding, Wang, Wu, and Zhang]{Yang_2025_ICCV}
Zhuoran Yang, Xi Guo, Chenjing Ding, Chiyu Wang, Wei Wu, and Yanyong Zhang.
\newblock Instadrive: Instance-aware driving world models for realistic and consistent video generation.
\newblock In \emph{Proceedings of the IEEE/CVF International Conference on Computer Vision (ICCV)}, pages 25410--25420, 2025{\natexlab{a}}.

\bibitem[Yang et~al.(2025{\natexlab{b}})Yang, Teng, Zheng, Ding, Huang, Xu, Yang, Hong, Zhang, Feng, Yin, Zhang, Wang, Cheng, Bin, Gu, Dong, and Tang]{ICLR2025_ce31378e}
Zhuoyi Yang, Jiayan Teng, Wendi Zheng, Ming Ding, Shiyu Huang, Jiazheng Xu, Yuanming Yang, Wenyi Hong, Xiaohan Zhang, Guanyu Feng, Da Yin, Yuxuan Zhang, Weihan Wang, Yean Cheng, Xu Bin, Xiaotao Gu, Yuxiao Dong, and Jie Tang.
\newblock Cogvideox: Text-to-video diffusion models with an expert transformer.
\newblock In \emph{International Conference on Learning Representations}, pages 83048--83077, 2025{\natexlab{b}}.

\bibitem[Yang et~al.(2026)Yang, Liu, Lu, Hou, Miao, Peng, Feng, Bai, and Zhao]{Yang_2026_CVPR}
Zhenya Yang, Zhe Liu, Yuxiang Lu, Liping Hou, Chenxuan Miao, Siyi Peng, Bailan Feng, Xiang Bai, and Hengshuang Zhao.
\newblock Geniedrive: Towards physics-aware driving world model with 4d occupancy guided video generation.
\newblock In \emph{Proceedings of the IEEE/CVF Conference on Computer Vision and Pattern Recognition (CVPR)}, pages 35680--35690, 2026.

\bibitem[Yuan et~al.(2024)Yuan, Zhang, Wang, Wei, Feng, Pan, Zhang, Liu, Albanie, and Ni]{Yuan_2024_CVPR}
Hangjie Yuan, Shiwei Zhang, Xiang Wang, Yujie Wei, Tao Feng, Yining Pan, Yingya Zhang, Ziwei Liu, Samuel Albanie, and Dong Ni.
\newblock Instructvideo: Instructing video diffusion models with human feedback.
\newblock In \emph{Proceedings of the IEEE/CVF Conference on Computer Vision and Pattern Recognition (CVPR)}, pages 6463--6474, 2024.

\bibitem[Zeng et~al.(2026)Zeng, Wu, Xiong, Wei, Guo, Zhu, Ho, Zhou, Zeng, Lu, Sun, Wang, Chen, Ye, and Zhang]{ICLR2026_db6caae0}
Kai Zeng, Zhanqian Wu, Kaixin Xiong, Xiaobao Wei, Xiangyu Guo, Zhenxin Zhu, Kalok Ho, Lijun Zhou, Bohan Zeng, Ming Lu, Haiyang Sun, Bing Wang, Guang Chen, Hangjun Ye, and Wentao Zhang.
\newblock Rethinking driving world model as synthetic data generator for perception tasks.
\newblock In \emph{International Conference on Learning Representations}, pages 135507--135534, 2026.

\bibitem[Zhang et~al.(2026)Zhang, Wu, Chen, Ji, Xiao, Huang, and Han]{Zhang_2026_WACV}
Jiacheng Zhang, Jie Wu, Weifeng Chen, Yatai Ji, Xuefeng Xiao, Weilin Huang, and Kai Han.
\newblock Align video diffusion model with online video-centric preference optimization.
\newblock In \emph{Proceedings of the IEEE/CVF Winter Conference on Applications of Computer Vision (WACV)}, pages 6142--6152, 2026.

\bibitem[Zhao et~al.(2023)Zhao, Gu, Varma, Luo, Huang, Xu, Wright, Shojanazeri, Ott, Shleifer, Desmaison, Balioglu, Damania, Nguyen, Chauhan, Hao, Mathews, and Li]{Zhao_2023_FSDP}
Yanli Zhao, Andrew Gu, Rohan Varma, Liang Luo, Chien-Chin Huang, Min Xu, Less Wright, Hamid Shojanazeri, Myle Ott, Sam Shleifer, Alban Desmaison, Can Balioglu, Pritam Damania, Bernard Nguyen, Geeta Chauhan, Yuchen Hao, Ajit Mathews, and Shen Li.
\newblock {PyTorch FSDP}: Experiences on scaling fully sharded data parallel.
\newblock \emph{Proceedings of the VLDB Endowment}, 16\penalty0 (12):\penalty0 3848--3860, 2023.

\bibitem[Zhou and Kr\"ahenb\"uhl(2022)]{Zhou_2022_CVPR}
Brady Zhou and Philipp Kr\"ahenb\"uhl.
\newblock Cross-view transformers for real-time map-view semantic segmentation.
\newblock In \emph{Proceedings of the IEEE/CVF Conference on Computer Vision and Pattern Recognition (CVPR)}, pages 13760--13769, 2022.

\bibitem[Zhou et~al.(2026)Zhou, Shao, Wang, Zong, Li, and Waslander]{ICLR2026_a940e52c}
Yang Zhou, Hao Shao, Letian Wang, Zhuofan Zong, Hongsheng Li, and Steven Waslander.
\newblock Drivinggen: A comprehensive benchmark for generative video world models in autonomous driving.
\newblock In \emph{International Conference on Learning Representations}, pages 103502--103524, 2026.

\end{thebibliography}
